\documentclass{article}

\PassOptionsToPackage{numbers}{natbib}

\usepackage[final]{neurips_2025}

\usepackage{graphicx}
\usepackage[utf8]{inputenc} %
\usepackage[T1]{fontenc}    %
\usepackage{hyperref}       %
\usepackage{url}            %
\usepackage{booktabs}       %
\usepackage{adjustbox}
\usepackage{amsfonts}       %
\usepackage{nicefrac}       %
\usepackage{xspace}          %
\usepackage{amsmath}
\usepackage{cleveref}      %
\usepackage[table]{xcolor}   %
\usepackage{booktabs}        %
\usepackage{fontawesome}

\usepackage[table]{xcolor}    %
\usepackage{booktabs}         %
\usepackage{makecell}         %
\usepackage{tikz}             %
\usepackage{multirow}
\usepackage{subcaption}
\usepackage{makecell}

\definecolor{poseblue}{RGB}{100,176,255}    %
\definecolor{poseyellow}{RGB}{255,180,0}    %
\definecolor{poseorange}{RGB}{255,140,60}   %
\definecolor{posered}{RGB}{255,90,90}       %
\definecolor{posepurple}{RGB}{180,90,255}   %

\definecolor{taggray}{gray}{0.2}             %
\definecolor{tagpurple}{RGB}{230,204,255}    %
\definecolor{tagnavy}{RGB}{25,25,112}        %
\definecolor{tagforest}{RGB}{34,139,34}      %
\definecolor{tagmaroon}{RGB}{128,0,0}        %
\definecolor{tagteal}{RGB}{0,128,128}        %

\newcommand{\best}[1]{\cellcolor{blue!20}\strut #1}
\newcommand{\second}[1]{\cellcolor{orange!20}\strut #1}

\usepackage{subcaption}

\usepackage{graphicx}
\usepackage[utf8]{inputenc} %
\usepackage[T1]{fontenc}    %
\usepackage{hyperref}       %
\usepackage{url}            %
\usepackage{booktabs}       %
\usepackage{amsfonts}       %
\usepackage{nicefrac}       %
\usepackage{microtype}      %
\usepackage{xcolor}         %
\usepackage{booktabs}       %
\usepackage{xspace}          %
\usepackage{amsmath}
\usepackage{cleveref}      %

\usepackage{algorithm}
\usepackage{algpseudocode}
\usepackage{placeins}
\usepackage{fontawesome}

\hypersetup{
    colorlinks=true,
    linkcolor={blue!60!black},        %
    citecolor={blue!60!black},        %
    urlcolor={red!65!black},          %
    filecolor={red!65!black},
    breaklinks=true
}

\newcommand{\ie}{\textit{i.e.}\@\xspace}
\newcommand\nnfootnote[1]{%
  \begin{NoHyper}
  \renewcommand\thefootnote{}\footnote{#1}%
  \addtocounter{footnote}{-1}%
  \end{NoHyper}
}

\newcommand{\oursMSF}{SegMASt3R}
\newcommand{\master}{MASt3R}

\title{\oursMSF: Geometry Grounded Segment Matching}

\author{%
\textbf{Rohit Jayanti}$^{1 * \dagger}$ \qquad
\textbf{Swayam Agrawal}$^{1 * \ddagger}$ \qquad
\textbf{Vansh Garg}$^{1 *}$ \qquad
\textbf{Siddharth Tourani}$^{2, 3}$ \\[0.8ex]
\textbf{Muhammad Haris Khan}$^3$ \qquad
\textbf{Sourav Garg}$^4$  \qquad
\textbf{Madhava Krishna}$^1$\\[1.6ex]
$^{1}$IIIT Hyderabad \qquad
$^{2}$University of Heidelberg \qquad
$^{3}$MBZUAI\qquad
$^{4}$Independent\\[1.6ex]
\faGithub\;Project Page: \url{https://segmast3r.github.io/}
}

\begin{document}

\maketitle

\nnfootnote{$^*$ Equal Contribution. $^\dagger$ Corresponding Author. $^\ddagger$ Now at Google DeepMind.}

\begin{abstract}
Segment matching is an important intermediate task in computer vision that establishes correspondences between semantically or geometrically coherent regions across images. Unlike keypoint matching, which focuses on localized features, segment matching captures structured regions, offering greater robustness to occlusions, lighting variations, and viewpoint changes. In this paper, we leverage the spatial understanding of 3D foundation models to tackle wide-baseline segment matching, a challenging setting involving extreme viewpoint shifts. We propose an architecture that uses the inductive bias of these 3D foundation models to match segments across image pairs with up to $180^\circ$ rotation. Extensive experiments show that our approach outperforms state-of-the-art methods, including the SAM2 video propagator and local feature matching methods, by up to 30\% on the AUPRC metric, on ScanNet++ and Replica datasets. We further demonstrate benefits of the proposed model on relevant downstream tasks, including 3D instance mapping and object-relative navigation.
\end{abstract}

\section{Introduction}

Segment matching establishes correspondences between coherent regions---objects, parts, or semantic segments---across images. It underpins video object tracking~\cite{conceptgraph}, scene-graph construction~\cite{hughes2022hydra, li2022sgtr}, robot navigation~\cite{garg2025objectreact, RoboHop, podgorski2025tango}, and instance-level SLAM~\cite{liang2023dig, xu2019mid}. Because it matches extended structures rather than sparse, texture-sensitive points, it is more robust to noise, occlusion, and appearance change. Mapping structured regions instead of isolated pixels also boosts interpretability and allows geometric or semantic priors to be injected, enabling higher-level reasoning about scene content.

Segment matching degrades sharply under \emph{wide‐baseline} conditions, where images of the same scene are taken from widely separated viewpoints that introduce severe perspective, scale, and up to $180^{\circ}$ rotation changes~\cite{ravi2024sam}.  Such cases arise in long-term video correspondence and robotic navigation, and demand models that reason over global 3D structure and enforce geometric consistency.  Current approaches that depend on features from pre-trained encoders like DINOv2~\cite{oquab2023dinov2} or ViT~\cite{dosovitskiy2020image} often mismatch repetitive patterns or fail to link drastically different views of the same object.

In this paper, we propose to leverage the strong spatial inductive bias of a 3D foundation model (namely MASt3R~\cite{leroy2024grounding}) to solve the problem of wide-baseline segment matching. 3D Foundation Models (3DFMs) are large-scale vision models trained to capture spatial and structural properties of scenes, such as depth, shape, and pose. Unlike appearance-focused models, 3DFMs learn geometry-aware representations using 2D and 3D supervision, enabling strong generalization across tasks like reconstruction and pose estimation. Their inductive bias toward spatial reasoning makes them well-suited for applications requiring geometric consistency such as our problem of wide-baseline segment matching.

We adapt \master{} for segment matching by appending a lightweight \emph{segment-feature head} that transforms patch-level embeddings into segment-level descriptors. Given an image pair, these descriptors are matched to establish segment correspondences. The head is trained with a contrastive objective patterned after SuperGlue~\cite{sarlin2020superglue}. Experiments show that our approach surpasses strong baselines, including SAM2’s video propagator which is trained on far larger datasets and state-of-the-art local feature matching methods. Finally, we demonstrate its practical utility in two downstream tasks: object-relative navigation and instance-level mapping.

\noindent\paragraph{Contributions} Our key contributions are summarized below:
\begin{itemize}
    \item We introduce a simple but effective approach for learning segment-aligned features by leveraging strong priors from a 3D foundation model (3DFM) \master. A differentiable segment matching layer is employed to align features across views, while a \emph{segment-feature head} transforms dense pixel-level representations into robust segment-level descriptors.
    
    \item To address the under-explored problem of wide-baseline segment matching, we construct a comprehensive benchmark comprising both direct segment association methods and those based on local feature matching. Our method demonstrates significant improvements over all baselines on challenging wide-baseline image pairs.
    
    \item We validate the practical utility of our approach by applying it to the downstream applications such as 3D instance mapping and object-relative topological navigation. Our method outperforms competitors by significant margins showcasing the efficacy of our design choices.
\end{itemize}

\section{Related Work}

\paragraph{Segment Matching and Segmentation Foundation Models.}
Robust segment-level association has emerged as a crucial intermediate step for high-level vision tasks such as scene graph construction, long-term object tracking, and multi-view instance association.  While related sub-problems like video instance segmentation and object tracking have been studied extensively in recent works~\cite{yan2023universal,yan2022towards,wang2020towards,edstedt2023dkm,pang2021quasi,meinhardt2022trackformer,li2022tracking,cetintas2023unifying}, the broader challenge of matching segments across arbitrary viewpoints, modalities, and time remains comparatively under-explored. Large-scale class-agnostic segmentation models have begun to close this gap.  The Segment Anything Model (SAM)~\cite{kirillov2023segment} and its successor SAM2~\cite{ravi2024sam} deliver high-quality masks and include a built-in propagator for associating masks across video frames.  However, this propagation module is optimized for short temporal windows and does not explicitly enforce geometric consistency under wide baselines or substantial appearance changes.

\paragraph{Learning to Match Segments and Overlap Prediction.}
Some recent methods address segment matching more directly.  MASA~\cite{li2024matching} augments SAM’s rich object proposals with synthetic geometric transformations to learn instance correspondences, and DMESA~\cite{zhang2024dmesa} extends these ideas to dense matching with improved efficiency. Despite such progress, these approaches are still limited by 2D supervision.
An alternative line of work predicts the degree of visual overlap between images~\cite{chen2022guide,fu2022learning,arnold2022mapfree}.  By estimating shared content, these methods implicitly learn region correspondences, yet they also remain confined to 2D training signals.

\paragraph{Local Feature Matching.}
Sparse~\cite{detone18superpoint,sarlin2020superglue,sun2021loftr,lindenberger2023lightglue,xuelun2024gim} and dense~\cite{edstedt2024roma,berton2021viewpoint} local feature matchers propagate pixel-level correspondences that can, in principle, transfer segment labels between views~\cite{RoboHop, garg2025objectreact}.  Nonetheless, like the previous categories, they are trained exclusively on image data and struggle with extreme viewpoint changes.

Across segment matching, overlap prediction, and feature matching, reliance on purely 2D supervision leaves existing techniques brittle under wide-baseline conditions.  Our method addresses this limitation by fine-tuning the 3D foundation model \mbox{MASt3R}~\cite{leroy2024grounding}, whose strong geometric priors enable reliable segment correspondence even when image pairs differ by nearly $180^{\circ}$ in viewpoint.

\begin{figure}
    \centering
    \includegraphics[width=1\linewidth]{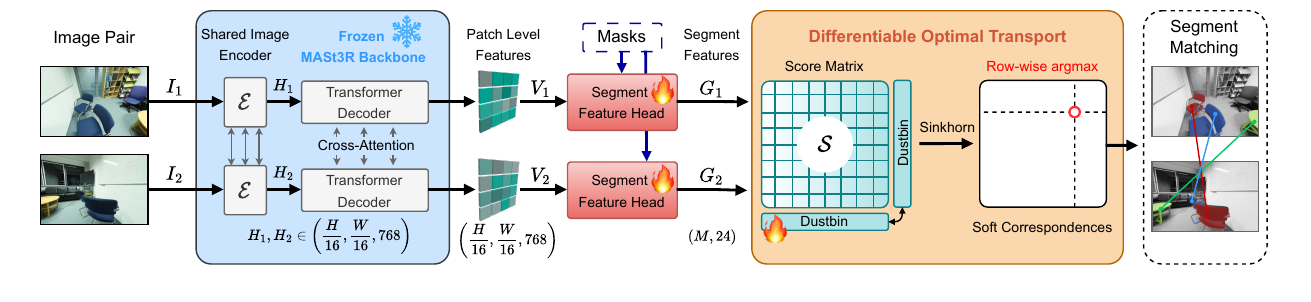}
    \caption{\textbf{Pipeline Overview}: An image pair is processed by a frozen MASt3R backbone to extract patch‑level features; segmentation masks are obtained either from a parallel segmentation module or ground truth annotation; the patch-level features are aggregated by the segment-feature head to form segment-level descriptors; and these descriptors are then matched across images via a differentiable optimal transport layer to produce segment-level correspondences.}
    \label{fig:overview}
\end{figure}

\vspace{-1em}

\section{Method}
\Cref{fig:overview} provides an overview of our method. It builds upon the \master~\cite{leroy2024grounding} architecture by introducing a Feat2Seg Adapter that maps the patch-level features output by the \master~decoder to get segment-level features, which are subsequently matched via a differentiable optimal transport and a row-wise argmax yielding the final segment-level correspondences. 

\subsection{MASt3R Preliminaries} MASt3R is a 3D foundation model pre‑trained on a diverse collection of 3D‑vision datasets commonly used for tasks such as metric depth estimation and camera‑pose prediction. Given a pair of images, it produces dense 3D point maps for each image and identifies correspondences between them. Thanks to training on data with wide‑baseline pairs, MASt3R generalizes well to unseen image pairs~\cite{leroy2024grounding} and consistently outperforms alternative 3D matching methods—such as MicKey~\cite{barroso2024matching},  which relies on a DINOv2 backbone~\cite{barroso2024matching}.

\noindent\paragraph{Architecture:} We summarize here the portions of the architecture we utilize in our pipeline. The two images \(I_1\) and \(I_2\) are processed in a Siamese manner by a weight‑sharing ViT encoder~\cite{dosovitskiy2020image} $\mathcal{E}$, resulting in token sets $H_1$, $H_2$, \ie
\[H_1 = \mathcal{E}(I_1),  \qquad H_2 = \mathcal{E}(I_2). \]

\paragraph{Cross-view transformer decoder:} Next, a pair of CroCo~\cite{weinzaepfel2022croco, weinzaepfel2023croco} style intertwined transformer decoders jointly refines the two feature sets.   By alternating self‑ and cross‑attention, the decoders exchange information between inputs to capture both the relative viewpoints and the global 3‑D structure of the scene.  We show via an ablation study, these cross-view aware decoders provide a significant boost to the segment matching in~\Cref{subsec:qual-results}

The resulting geometry‑aware representations are denoted \(V_{1}\) and \(V_{2}\):
\[
\bigl({V}_{1},\;{V}_{2}\bigr)
  = \operatorname{Decoder}\bigl(\mathbf{H}_{1},\;\mathbf{H}_{2}\bigr).
\]
Assuming, $H, W$ denote the height and width of the input images, the output geometry‑aware patch-level features are of size $(H/16,W/16,768)$. The original architecture subsequently uses two prediction heads to output dense point-maps, as well as pixel-level features, they are not shown in~\Cref{fig:overview} as they are not used in our pipeline. These portions of the pipeline remain frozen and their outputs are used as is, we instead introduce a new segment-feature head.

\subsection{Segment-Feature Head: Segment-Aligned Features}
\label{subsec:feat2seg}

The MASt3R decoder produces patch--level embeddings
$V_1,V_2$ of shape $(H/16, W/16,768)$.
In the original MASt3R~\cite{leroy2024grounding}, a
\emph{feature head} upsamples these tensors to the resolution of the input
image.  For \emph{segment matching} we introduce another head to transform the patch level features to $M$ segment features. This introduced head is realized as an MLP that upsamples the patch-level features $(V_1,V_2)$ to image resolution, yielding feature maps of size $(H,W,24)$, $24$ being the feature dimension. We denote this \textit{Feature-to-Segment} head as the \textbf{segment-feature} head. In addition to the patch-features, the \textbf{segment-feature} takes as input $M$ image resolution segment masks, obtained either from an external segmenter such as SAM2~\cite{ravi2024sam} or from ground-truth
annotations. Both the masks and feature maps are flattened along the spatial dimensions yielding flattened tensors.

\[
\mathbf{P}_{\mathrm{flat}}\in\mathbb{R}^{24\times HW},\qquad
\mathbf{M}_{\mathrm{flat}}\in\mathbb{R}^{M\times HW}.
\]
To go from pixel-level descriptors to segment descriptors, a single batched matrix multiplication aggregates the pixel descriptors inside each mask:
\begin{equation}
\mathbf{G}= \mathbf{M}_{\mathrm{flat}}\;\mathbf{P}_{\mathrm{flat}}^{\!\top}
\in\mathbb{R}^{M\times 24}.
\label{eq:seg_feats}
\end{equation}
The resulting segment embeddings for the two images are denoted
$\mathbf{G}_1$ and $\mathbf{G}_2$ in \Cref{fig:overview} and are fed to the
differentiable matching layer described in
Section~\ref{subsec:diff_match}.  
We set $M=100$ as an upper bound for batch processing. In practice, images typically contain 20-30 masks when training with ground truth annotations; we pad with zeros when fewer masks are present. At inference time, the number of masks can be set arbitrarily, independent of the training-time value of $M$.

\subsection{Differentiable Segment Matching Layer}
\label{subsec:diff_match}
The goal of the differentiable segment matching layer is to establish permutation-style correspondences between the $M$ segments from each image, given segment descriptors $(G_1, G_2) \in \mathbb{R}^{(M,24)}$

\paragraph{Cosine–similarity affinity.}
We first construct an affinity matrix
$\mathbf{S}\in\mathbb{R}^{M_1\times M_2}$ with a simple dot product
\begin{equation}
S_{ij}= \langle\mathbf{g}^1_i,\mathbf{g}^2_j\rangle,
\quad
1\le i\le M_1,\;1\le j\le M_2.
\label{eq:cos_sim}
\end{equation}

$\mathbf{g}^1_i$ and $\mathbf{g}^2_j$ correspond to segment-level features from $G_1$ and $G_2$ respectively. Ideally, segment features corresponding to the same underlying 3D region should have a high similarity score and dis-similar regions a correspondingly low score.

\paragraph{Learnable dustbin.}
Following \cite{sarlin2020superglue}, we incorporate a \emph{dustbin} row and column in the affinity matrix $\mathbf{S}$ to handle segments without correspondences in the other image, which is critical for wide-baseline matching.
We augment~\eqref{eq:cos_sim} by concatenating an additional row and column
initialized with a learnable logit $\alpha\in\mathbb{R}$, yielding $\tilde{\mathbf{S}}\in\mathbb{R}^{(M_1+1)\times(M_2+1)}$.
\[
\tilde{\mathbf{S}} =
\begin{bmatrix}
\mathbf{S}           & \alpha\mathbf{1}_{M_1}\\
\alpha\mathbf{1}_{M_2}^{\!\top} & \alpha
\end{bmatrix}\!,
\]

\paragraph{Soft Correspondences via Sinkhorn} 
The similarity logits are transformed into a soft assignment matrix
$\mathbf{P}$ by $T$ iterations of the Sinkhorn normalisation~\cite{sinkhorn1967concerning,sinkhorn1964relationship} in log-space:
\begin{align}
\mathbf{P}^{(0)} &\leftarrow \exp \!\bigl(\tilde{\mathbf{S}}/\tau\bigr),\nonumber\\
u^{(t)}_i        &=\frac{1}{\sum_j P^{(t)}_{ij}},\quad
v^{(t)}_j=\frac{1}{\sum_i P^{(t)}_{ij}},\label{eq:sinkhorn}\\
P^{(t+1)}_{ij}   &=u^{(t)}_i\,P^{(t)}_{ij}\,v^{(t)}_j,\nonumber
\quad 0\le t<T, 
\end{align}
where $\tau$ is a temperature hyper-parameter.
After convergence,
$\mathbf{P}\!=\!\mathbf{P}^{(T)}$ is (approximately) doubly stochastic. Throughout all experiments conducted, we assume $T=50$. The output of the Sinkhorn algorithm $\mathbf{P}^{(T)}$ is a soft bi-stochastic matrix, which has to be discretized to obtain the final segment matches.

\paragraph{Discrete correspondences.} To obtain the simple final segment matches, we perform a simple row-wise $\arg\max$ over the \emph{non-dustbin}
columns:
\[
m(i)=\operatorname*{argmax}_{1\le j\le M_2} P_{ij},\quad
\text{with assignment accepted if } j\neq M+1.
\]

\subsection{Supervision}
\paragraph{Training objective.}
We adopt the SuperGlue cross-entropy loss
$\mathcal{L}_{\text{SG}}$~\cite{sarlin2020superglue}, extended with explicit
terms for unmatched segments:
\begin{equation}
\mathcal{L}
= -\!\!\sum_{(i,j)\in\mathcal{M}}
        \log P_{ij}
  -\!\!\sum_{i\in\mathcal{U}_1}\!\!\log P_{i,M+1}
  -\!\!\sum_{j\in\mathcal{U}_2}\!\!\log P_{M+1,j},
\label{eq:loss}
\end{equation}
where $\mathcal{M}$ is the set of ground-truth matches and
$\mathcal{U}_1,\mathcal{U}_2$ are the unmatched indices in
image~1 and~2, respectively.  The dustbin parameter $\alpha$ is learned
jointly with the rest of the network, enabling the layer to balance match
confidence against the cost of declaring non-matches. This fully differentiable design allows the matching layer to be trained end-to-end together with the upstream segment encoders and downstream task losses.

\subsubsection{Training Details}
\label{subsubsec:training-details}

The model is trained using AdamW optimizer with an initial learning rate of \texttt{1e-4}, weight decay of \texttt{1e-4}, and a cosine annealing learning rate schedule without restarts, decaying up to a minimum learning rate of \texttt{1e-6} over the full training duration. We use a batch size of 36 and train the model for 20 epochs on a single NVIDIA RTX A6000 GPU.  The \textbf{segment-feature heads} are initialized with MASt3R's local feature head weights and finetuned further. For the differentiable segment matcher we initialize the single learnable dustbin parameter to $1.0$. The number of Sinkhorn iterations is set to 50. Training our model on ScanNet++ takes 22 hours, whereas a single forward pass during inference with batch size of 1 takes 0.579 seconds.

\vspace{-1em}
\section{Experiments}
\label{sec:experiments}

\subsection{Datasets}
\label{subsec:datasets}
Our network is trained on scenes from ScanNet++~\cite{yeshwanthliu2023scannetpp} which contain a diverse set of real-world scenes, primarily in indoor settings. We test our model on novel scenes from ScanNet++ as well as perform  cross-dataset generalization studies on Replica~\cite{replica19arxiv} and MapFree~\cite{arnold2022mapfree} datasets. The former  contains high-quality photo-realistic indoor scenes while the latter is a challenging outdoor visual-localization dataset, which is sufficiently out-of-distribution considering our training data. %

\paragraph{ScanNet++~\cite{yeshwanthliu2023scannetpp}.} ScanNet++ contains 1\,006 indoor scenes captured with DSLR images and RGB-D iPhone streams, all co-registered to high-quality laser scans.  The dataset supplies 3D semantic meshes, 2D instance masks, and accurate camera poses, which we utilize in our pipeline. We train on 860\,k image pairs from 140 scenes and evaluate on 8\,k pairs from 36 validation scenes, sampled with a fixed seed (42) and balanced across scenes and four pose bins: $[0^\circ\!-\!45^\circ]$, $[45^\circ\!-\!90^\circ]$, $[90^\circ\!-\!135^\circ]$, and $[135^\circ\!-\!180^\circ]$, defined by the rotational geodesic distance between camera orientations.

\paragraph{Replica~\cite{replica19arxiv}.} The Replica dataset contain 18 high-quality, photo-realistic indoor room reconstructions in the form of dense meshes replete with per-primitive semantic class and instance information. In particular, we use a pre-rendered version of this dataset from Semantic-NeRF~\cite{Zhi:etal:ICCV2021}, which directly provides poses, RGB-D sequences, and per-frame semantic masks.  For evaluation, we employ the same pose-binning strategy as described above and randomly sample 3200 image pairs across 8 scenes, again ensuring uniform sampling across both scenes and pose-bins.

\paragraph{MapFree Visual Re-localization~\cite{arnold2022mapfree}.} MapFree is a challenging benchmark for metric-relative pose estimation, featuring 655 diverse outdoor scenes (sculptures, fountains, murals) with extreme viewpoint changes, varying visual conditions, and geometric ambiguities. The training split contains 460 scenes with 0.5M images; we uniformly sample 50 scenes yielding 31K image pairs for training. Since test split ground truth is unavailable, we report results on the validation split using 7.8K pairs from 13 uniformly sampled scenes (out of 65 total).

\subsection{Baselines}
\label{subsec:baselines}
\paragraph{Local Feature Matching (LFM).} Although few models target \emph{segment} matching directly, a wide range of local feature matchers (LFMs) exists at different densities: sparse SuperPoint~\cite{detone18superpoint}, semi-dense LoFTR~\cite{sun2021loftr}, and dense RoMa and MASt3R~\cite{edstedt2024roma,leroy2024grounding}.  
We harness these state-of-the-art LFMs via the EarthMatch toolkit~\cite{Berton_2024_EarthMatch} to obtain segment correspondences.  
Each matched keypoint pair votes for the source and target masks that contain its coordinates, populating a vote matrix of size $M\times N$ (with $M$ and $N$ segments in the two images).  Correspondences are taken as the highest-scoring entries of this matrix; the full algorithm is provided in the supplementary.

\paragraph{Segment Matching (SegMatch).}  We also compute segment matches from dense features of two strong pre-trained vision encoders, DINOv2~\cite{oquab2023dinov2} and MASt3R~\cite{leroy2024grounding}, as well as SAM2's video propagator~\cite{ravi2024sam} to track masks across views.  For feature based methods, masks are downsampled via nearest-neighbor interpolation to match feature resolution, and segment descriptors are computed via masked average pooling. Cosine similarity between descriptors yields a match matrix, from which one-to-one links are selected via mutual-check. We further benchmark against SegVLAD~\cite{garg2024revisitanything}, which aggregates features from neighbouring segments for segment retrieval based visual place recognition.

\subsection{Evaluation Metrics}
\label{subsec:metrics} We report two complementary measures of segment correspondence quality - \textbf{AUPRC} and \textbf{Recall}. 
\emph{Area Under the Precision--Recall Curve (AUPRC)} integrates precision over the entire recall axis,
providing a threshold–independent summary that is particularly informative under the high class–imbalance 
characteristic of segment matching between image pairs. 
\emph{Recall@\(k\)} (\(R@k\)) denotes the fraction of query segments whose ground-truth counterpart is found 
within the top \(k\) ranked candidates, thus gauging how effectively the method surfaces correct matches among 
its highest-confidence predictions.
{Additional details for the dataset, baseline and experiments can be found in the supplementary, along with more qualitative results.}

\section{Results}\label{subsec:qual-results}

\begin{table*}[t]
\centering
\small
\resizebox{\textwidth}{!}{%
\begin{tabular}{%
  >{\raggedright\arraybackslash}p{1.5cm}  %
  >{\raggedright\arraybackslash}p{2cm}  %
  ccc   %
  ccc   %
  ccc   %
  ccc}  %
\toprule
\textbf{Type}
 & \textbf{Method}
 & \multicolumn{3}{c}{\textbf{0\textdegree–45\textdegree}}
 & \multicolumn{3}{c}{\textbf{45\textdegree–90\textdegree}}
 & \multicolumn{3}{c}{\textbf{90\textdegree–135\textdegree}}
 & \multicolumn{3}{c}{\textbf{135\textdegree–180\textdegree}} \\
\cmidrule(lr){3-5} \cmidrule(lr){6-8} \cmidrule(lr){9-11} \cmidrule(lr){12-14}
 &  & AUPRC & R@1 & R@5
      & AUPRC & R@1 & R@5
      & AUPRC & R@1 & R@5
      & AUPRC & R@1 & R@5 \\
\midrule
\multirow{4}{2cm}{Local Feature\\Matching (LFM)}
 & SP-LG~\cite{detone18superpoint,lindenberger2023lightglue}
   & 42.1 & 45.6 & 51.2
   & 33.5 & 36.9 & 43.1
   & 15.9 & 19.7 & 26.2
   & 6.1  & 9.3  & 14.6 \\

 & GiM-DKM~\cite{xuelun2024gim,edstedt2023dkm}
   & 59.1 & 64.9 & 69.7
   & 54.9 & 60.2 & 66.1
   & 39.6 & 44.5 & 51.8
   & 21.3 & 25.9 & 32.7 \\

 & RoMA~\cite{edstedt2024roma}
   & 61.6 & \second{68.7} & 73.5
   & \second{58.9} & \second{66.4} & 73.0
   & 47.4 & 56.1 & 65.5
   & 30.0 & 39.5 & 49.7 \\

 & MASt3R~\cite{leroy2024grounding}
   & 59.5 & 68.3 & 74.2
   & 57.3 & 65.6 & 72.5
   & \second{52.9} & \second{60.3} & 68.9
   & \second{45.4} & \second{52.6} & 62.2 \\
\midrule
\multirow{4}{2cm}{Segment Matching\\(SegMatch)}
 & SAM2~\cite{ravi2024sam}
   & \second{61.9} & 64.6 & 67.5
   & 46.6 & 50.1 & 54.0
   & 27.9 & 32.5 & 37.2
   & 17.0 & 21.6 & 25.4 \\

 & DINOv2~\cite{oquab2023dinov2}
   & 57.9 & 66.7 & \second{87.4}
   & 43.0 & 55.9 & \second{83.2}
   & 33.5 & 48.0 & \second{78.0}
   & 32.4 & 46.0 & \second{75.6} \\

 & SegVLAD~\cite{garg2024revisitanything}
   & 44.2 & 58.6 & 81.4
   & 32.1 & 49.5 & 76.5
   & 23.2 & 42.2 & 70.5
   & 20.0 & 39.6 & 66.8 \\

 & MASt3R~\cite{leroy2024grounding}
   & 51.7 & 54.6 & 69.9
   & 45.6 & 49.8 & 68.5
   & 41.4 & 47.9 & 69.2
   & 39.5 & 48.7 & 72.6 \\
\midrule
Ours
 & \textsc{\oursMSF{}}
   & \best{92.8} & \best{93.6} & \best{98.0}
   & \best{91.1} & \best{92.2} & \best{97.6}
   & \best{88.0} & \best{89.5} & \best{96.8}
   & \best{83.6} & \best{85.9} & \best{95.9} \\
\bottomrule
\end{tabular}%
} %
\caption{Performance of selected methods across pose‐bins on ScanNet++~\cite{yeshwanthliu2023scannetpp}. \colorbox{blue!20}{Blue} cells mark the best scores; \colorbox{orange!20}{Orange} cells mark the second‐best.}
\label{tab:segmatch_posebins_scannetpp}
\end{table*}

\paragraph{Segment Matching}
In Table~\ref{tab:segmatch_posebins_scannetpp}, we compare our proposed method~\oursMSF{} with the state-of-the-art methods using two categories of approaches in the literature: the well-established local feature matching (LFM) based on sparse or dense keypoint correspondences, and the recently emerging open-set instance association based on segment matching (SegMatch). It can be observed that~\oursMSF{}~outperforms all the baselines for all the pose bins with a huge margin. Amongst the LFM techniques, dense matchers (RoMa and \master{}) outperform sparse matchers (SP-LG and GiM-DKM) on both the evaluation metrics for the task of segment matching. Notably, on the highly challenging wide-baseline settings, \master{} outperforms other local matchers by a large margin. However, when the same backend features are aggregated at the segment-level, \master{}'s performance deteriorates significantly, e.g., AUPRC drops from 52.9 to 41.4 on the 90-135 pose bin. This uncovers an inherent limitation: \textit{feature distinctiveness at the pixel level does not necessarily translate to the instance level}. Considering the SegMatch techniques which are not trained for the LFM task, it can be observed that DINOv2 and SegVLAD perform particulary well on R@5 metric, which aligns with their typical use for coarse retrieval~\cite{anyloc,garg2024revisitanything}. On the other hand, SAM2's two-frame video propagation only works well for narrow-baseline matching, which can be expected as its training set is mostly comprised of dynamic object tracking. Overall, these results show that all the baseline methods lack on at least one of the fronts: pixel- vs segment-level distinctiveness, recall vs precision, and narrow- vs wide-baseline robustness. Our proposed method~\oursMSF{} achieves all these desirable properties with high performance across the board by learning segment-level representations with the training objective of instance association. In Section~\ref{subsec:qual-results}, we provide qualitative results which emphasize the ability of our method to address the problem of perceptual instance aliasing and instance matching under extreme viewpoint shifts. 

\vspace{-0.5em}

\paragraph{Generalization}
We assess our model's ability to generalize to new environments. First, in~\Cref{tab:segmatch_posebins_replica}, we present results on a different indoor dataset, Replica, to test the generalization ability of \oursMSF{}, which is trained only on ScanNet++.  We compare \oursMSF{} against the LFM and SegMatch versions of \master, as these three methods closely resemble each other in terms of their network architecture (detailed comparisons are included in the supplementary). It can be observed that the performance patterns on Replica remain largely the same as that on ScanNet++, and \oursMSF{} consistently outperforms the LFM and SegMatch versions of \master{} across the board.

Furthermore, we test generalization to challenging outdoor scenes from the MapFree dataset~\cite{arnold2022mapfree}, with results shown in~\Cref{tab:segmatch_mapfree}. Since MapFree lacks instance-level ground truth, we use SAM2's video propagator on image sequences to generate a pseudo-ground truth for evaluation. Our indoor-trained model (\oursMSF{} (SPP)) shows a regression in performance in comparison to DINOv2, highlighting the domain shift. However, this gap can be substantially closed by either re-training on MapFree data (\oursMSF{} (MF)) or, even more simply by just recalibrating the single learnable dustbin parameter $\alpha$ using a grid-search over a small calibration set from the target domain (\oursMSF{} (SPP, Dustbin MF)). This demonstrates the strong adaptability of our model's learned geometric features.

\begin{table*}[t]
\centering
\small
\resizebox{\textwidth}{!}{%
{\renewcommand{\arraystretch}{1.0}
\begin{tabular}{%
  >{\raggedright\arraybackslash}p{1cm}  %
  >{\raggedright\arraybackslash}p{2cm}    %
  ccc   %
  ccc   %
  ccc   %
  ccc}  %
\toprule
\textbf{Type}
 & \textbf{Method}
 & \multicolumn{3}{c}{\textbf{0\textdegree–45\textdegree}}
 & \multicolumn{3}{c}{\textbf{45\textdegree–90\textdegree}}
 & \multicolumn{3}{c}{\textbf{90\textdegree–135\textdegree}}
 & \multicolumn{3}{c}{\textbf{135\textdegree–180\textdegree}} \\
\cmidrule(lr){3-5} \cmidrule(lr){6-8} \cmidrule(lr){9-11} \cmidrule(lr){12-14}
 &  & AUPRC & R@1 & R@5
      & AUPRC & R@1 & R@5
      & AUPRC & R@1 & R@5
      & AUPRC & R@1 & R@5 \\
\midrule

 LFM & MASt3R~\cite{leroy2024grounding}
   & \second{78.2} & \second{86.5} & \second{89.4}
   & \second{69.5} & \second{77.6} & \second{81.0}
   & \second{48.0} & \second{60.4} & 64.6
   & \second{32.5} & \second{49.0} & 54.1 \\

 SegMatch & MASt3R~\cite{leroy2024grounding}
   & 52.2 & 57.5 & 81.2
   & 39.1 & 51.0 & 78.6
   & 23.6 & 45.9 & \second{77.2}
   & 17.2 & 43.8 & \second{75.7} \\

Ours
 & \textsc{SegMASt3R}
   & \best{95.0} & \best{96.0} & \best{98.6}
   & \best{86.2} & \best{91.2} & \best{96.4}
   & \best{73.4} & \best{85.2} & \best{95.7}
   & \best{68.4} & \best{83.8} & \best{94.8} \\

\bottomrule
\end{tabular}%
}}
\caption{Performance of selected methods across pose‐bins on Replica~\cite{replica19arxiv}. \colorbox{blue!20}{Blue} cells mark the best scores; \colorbox{orange!20}{Orange} cells mark the second‐best.}
\label{tab:segmatch_posebins_replica}
\end{table*}

\begin{table}[h]
\centering
\small
\resizebox{0.8\linewidth}{!}{%
\begin{tabular}{lcccccc}
\toprule
\textbf{Method} & \textbf{Train Set} & \textbf{Overall IoU} & \textbf{0-45\textdegree} & \textbf{45-90\textdegree} & \textbf{90-135\textdegree} & \textbf{135-180\textdegree} \\
\midrule
DINOv2~\cite{oquab2023dinov2} & Multiple & 84.4 & 85.4 & 85.2 & 83.5 & 83.8 \\
MASt3R (Vanilla)~\cite{leroy2024grounding} & Multiple & 69.2 & 73.4 & 69.8 & 70.0 & 66.1 \\
\midrule
\oursMSF{} (SPP) & ScanNet++ & 75.2 & 75.2 & 74.6 & 76.5 & 74.5 \\
\oursMSF{} (SPP, Dustbin MF) & ScanNet++ & 88.7 & 88.6 & 88.6 & 88.5 & 88.9 \\
\oursMSF{} (MF) & MapFree & \best{93.7} & \best{93.3} & \best{93.7} & \best{93.9} & \best{93.9} \\
\bottomrule
\end{tabular}%
} %
\vspace{0.5em}
\caption{Generalization performance on the outdoor MapFree dataset~\cite{arnold2022mapfree}.}

\label{tab:segmatch_mapfree}
\end{table}

\paragraph{3D Instance Mapping}

\begin{table}[htbp]
  \centering
  \setlength{\tabcolsep}{4pt}
  \renewcommand{\arraystretch}{1.2}
  \resizebox{\linewidth}{!}{%
\begin{tabular}{l*{8}{c}}
  \toprule
  \textbf{Method}
    & \textbf{office0}
    & \textbf{office1}
    & \textbf{office2}
    & \textbf{office3}
    & \textbf{office4}
    & \textbf{room0}
    & \textbf{room1}
    & \textbf{room2} \\
  \cmidrule(lr){2-2}
  \cmidrule(lr){3-3}
  \cmidrule(lr){4-4}
  \cmidrule(lr){5-5}
  \cmidrule(lr){6-6}
  \cmidrule(lr){7-7}
  \cmidrule(lr){8-8}
  \cmidrule(lr){9-9}
  
    & {AP / AP@50}
    & {AP / AP@50}
    & {AP / AP@50}
    & {AP / AP@50}
    & {AP / AP@50}
    & {AP / AP@50}
    & {AP / AP@50}
    & {AP / AP@50} \\
  \midrule
  ConceptGraphs (MobileSAM Masks)~\cite{conceptgraph}
    & 11.84 / 28.43
    & 20.31 / 43.79
    & 8.63 / 22.82
    & 8.07 / 22.83
    & 9.46 / 24.73
    & 12.23 / 34.34
    & 5.83 / 12.96
    & 7.83 / 23.82 \\
  ConceptGraphs (GT Masks)~\cite{conceptgraph}
    & 43.53 / 69.68
    & 22.48 / 40.71
    & 43.46 / 60.69
    & 32.06 / 53.44
    & 39.63 / 68.22
    & 44.89 / 69.64
    & 17.96 / 36.53
    & 25.93 / 43.63 \\
  \oursMSF (Ours, GT Masks)
    & \cellcolor{blue!15}79.93 / \cellcolor{blue!15}87.17
    & \cellcolor{blue!15}54.89 / \cellcolor{blue!15}64.42
    & \cellcolor{blue!15}64.00 / \cellcolor{blue!15}85.50
    & \cellcolor{blue!15}58.02 / \cellcolor{blue!15}79.93
    & \cellcolor{blue!15}67.48 / \cellcolor{blue!15}85.01
    & \cellcolor{blue!15}71.02 / \cellcolor{blue!15}91.22
    & \cellcolor{blue!15}64.09 / \cellcolor{blue!15}85.50
    & \cellcolor{blue!15}56.35 / \cellcolor{blue!15}76.66 \\
  \bottomrule
\end{tabular}
}
  \vspace{0.5em}
  \caption{Class-Agnostic instance-mapping performance (AP and AP@50) on Replica scenes, shown in percentage.  The best value in each column is highlighted in blue.}
  \label{tab:instance_mapping_replica_main}
\end{table}

\vspace{-0.5em}
The goal of \emph{instance mapping} is to localize object instances in 3D so a robot can distinguish objects of the same class in both image and metric space~\cite{conceptgraph,cliploc,xu2019mid}.  The main difficulty is preserving identities over long trajectories, especially when objects leave the camera’s view and later reappear from different angles. Our pipeline employs \textsc{SegMASt3R} to match object masks across image pairs.  Given ground-truth masks and sampled pairs, we extract mask features, solve a Sinkhorn assignment, and take the row-wise argmax to obtain tentative correspondences.  We then back-project each matched mask into 3D and drop links whose point-cloud IoU falls below 0.5, rejecting any match that fails this geometric check.  Details are provided in the supplementary. Table~\ref{tab:instance_mapping_replica_supplementary} shows percentage AP versus ConceptGraphs~\cite{conceptgraph}. We evaluate under two conditions: using masks generated by MobileSAM~\cite{mobile_sam} as in the original ConceptGraphs setup, and using ground-truth (GT) masks for a fairer comparison of the underlying matching capability. Our geometry-aware matching yields higher accuracy in both settings, particularly when objects exit and later re-enter the field of view, demonstrating robustness to both noisy masks and challenging viewpoints.

\paragraph{Robustness to Noisy Segmentation Masks}
To assess the practical viability of our approach in real-world scenarios where ground-truth masks are unavailable, we evaluated all methods on ScanNet++ using imperfect segmentations generated by FastSAM. A key challenge in this setting is that evaluation can conflate the performance of the segment matcher with the quality of the upstream Automatic Mask Generator (AMG). A low score may not distinguish between a matching failure and an AMG failure where predicted masks do not align with any ground-truth segment.

To decouple these factors and isolate the core matching performance, we adopt an evaluation protocol that does not penalize methods for AMG errors. Specifically, while all methods take noisy masks as input, the evaluation is performed only on the subset of ground-truth correspondences for which a valid match between predicted segments was possible. As shown in Table~\ref{tab:segmatch_noisy_masks_spp}, \oursMSF{} maintains a substantial performance margin over all baselines across different evaluation thresholds. This result confirms that the strong geometric priors learned by our method provide significant robustness, enabling superior matching even when conditioned on inconsistent and noisy segment inputs.

\begin{table*}[t]
\centering
\small
\resizebox{\textwidth}{!}{%
\begin{tabular}{%
  >{\raggedright\arraybackslash}p{4cm} %
  ccc %
  ccc %
  ccc} %
\toprule
\textbf{Method}
  & \multicolumn{3}{c}{\textbf{IoU > 0.25}}
  & \multicolumn{3}{c}{\textbf{IoU > 0.50}}
  & \multicolumn{3}{c}{\textbf{IoU > 0.75}} \\
\cmidrule(lr){2-4} \cmidrule(lr){5-7} \cmidrule(lr){8-10}
  & AUPRC & R@1 & R@5
  & AUPRC & R@1 & R@5
  & AUPRC & R@1 & R@5 \\
\midrule
SAM2~\cite{ravi2024sam}
  & 47.4 & 56.2 & 61.2
  & 50.2 & 58.7 & 62.6
  & \second{55.6} & 64.1 & 66.6 \\

DINOv2~\cite{oquab2023dinov2}
  & 39.0 & \second{60.0} & \second{89.2}
  & 44.4 & \second{65.0} & \second{92.0}
  & 54.9 & \second{73.1} & \second{94.0} \\

MASt3R (Vanilla)~\cite{leroy2024grounding}
  & \second{50.7} & 58.0 & 76.2
  & \second{52.4} & 59.5 & 77.0
  & 49.6 & 57.4 & 76.3 \\
\midrule
SegMASt3R (Ours)
  & \best{84.2} & \best{91.9} & \best{99.3}
  & \best{87.6} & \best{94.4} & \best{99.3}
  & \best{89.9} & \best{94.3} & \best{96.3} \\
\bottomrule
\end{tabular}%
} %
\vspace{0.5em}
\caption{Performance on ScanNet++~\cite{yeshwanthliu2023scannetpp} using noisy masks from FastSAM.}
\label{tab:segmatch_noisy_masks_spp}
\end{table*}

\vspace{-0.5em}

\paragraph{Object-level Topological Navigation}

Recent works~\cite{conceptgraph, RoboHop, garg2025objectreact} have explored the use of SAM2's open-set, semantically-meaningful instance segmentation for topological mapping and navigation. These methods rely on accurate segment-level association. To show the benefits of our proposed method on complex downstream tasks such as navigation, we considered an object topology-based mapping and navigation method, RoboHop~\cite{RoboHop}, and swapped its SuperPoint+LightGlue based segment matching with our proposed segment matcher. This segment matching is used by RoboHop's localizer to match each of the object instances in the query image with those in the sub-map images, which is then used to estimate the currently-viewed object's path to the goal and correspondingly obtain a control signal. We used the val set of the ImageInstanceNav~\cite{krantz2022instance} dataset, which comprises real-world indoor scenes from HM3Dv0.2~\cite{ramakrishnan2021hm3d}. We followed~\cite{podgorski2025tango} for creating map trajectories and evaluating navigation performance.

\begin{figure}
  \centering
  \begin{tabular}{cc}

  \includegraphics[width=0.48\linewidth]{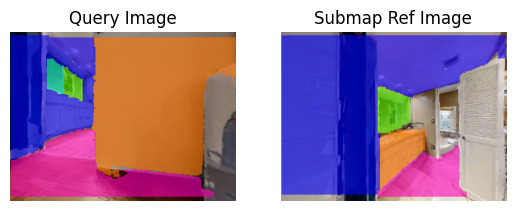} & 
 \includegraphics[width=0.48\linewidth]{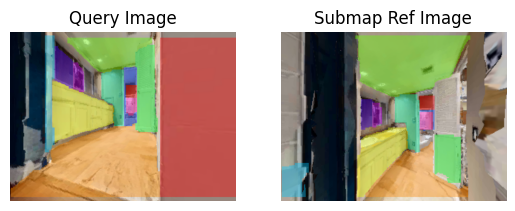} \\

  RoboHop w. DINOv2 & w. \oursMSF{} \\

  \end{tabular}

  \caption{\textbf{Segment Matching-Guided Navigation.} (left) In vanilla RoboHop's segment matching, a wall segment (orange) gets mismatched with the vanity cabinet and misguides the agent to move towards its right, leading to a navigation failure. (right) \oursMSF{} correctly recognizes the same cabinet as well as other segments and guides the robot into the bathroom, and eventually to the goal. Note that the query and submap images vary across both the methods, as we manually probed the point of failure for the baseline and the nearest agent state for ours.}
  \label{fig:nav_all}
\end{figure}

In Table~\ref{tab:navigation}, we report navigation success for (vanilla) RoboHop   and an enhanced version of it that uses~\oursMSF{} for segment association for localization and navigation. We use two metrics: SPL (Success weighted by Path Length)~\cite{anderson2018evaluation} and SSPL (soft SPL)~\cite{datta2021integrating}. Furthermore,  we considered four different evaluation settings using two parameters that define the submap used for localizing the query image segments: \textit{Submap Span} ($S_s$), which defines the total number of images sampled from the map based on the distance from the robot's current position, and \textit{Submap Density} ($S_\rho$), which defines a subsampling factor to uniformly skip map images. These parameter configurations aid in testing narrow- and wide-baseline matching as well as the ability to avoid false positives. \Cref{tab:navigation} shows that~\oursMSF{} consistently outperforms vanilla RoboHop. In particular, we achieve an absolute improvement of $27\%$ SPL on one of the hardest parameter settings: $S_s=16, S_\rho=0.25$, where only $4$ submap images are sampled due to a low density value. This shows that even with a very sparse submap, it is possible to maintain high navigation success rate, thus avoiding the typical trade-off between compute time and accuracy for the localizer. In Figure~\ref{fig:nav_all}, we present a qualitative comparison between vanilla RoboHop's segment matching and that based on \oursMSF{}: (left) the mismatch between the wall (orange) and the vanity cabinet for the vanilla methods leads to an incorrect rotation towards the right, consequently leading to navigation failure, whereas \oursMSF{}'s accurate segment matching (right) guides the agent into the bathroom and finally to the goal object.

\vspace{-1em}
\begin{table}[t]
  \centering
  \setlength{\tabcolsep}{5pt}
  \renewcommand{\arraystretch}{1.3}
  \resizebox{\linewidth}{!}{%
    \begin{tabular}{l
      cc  %
      cc  %
      cc  %
      cc  %
    }
      \toprule
      \textbf{Method}
        & \multicolumn{2}{c}{$S_s=16,\,S_\rho=0.25$}
        & \multicolumn{2}{c}{$S_s=16,\,S_\rho=0.5$}
        & \multicolumn{2}{c}{$S_s=32,\,S_\rho=0.25$}
        & \multicolumn{2}{c}{$S_s=32,\,S_\rho=0.5$} \\
      \cmidrule(lr){2-3}
      \cmidrule(lr){4-5}
      \cmidrule(lr){6-7}
      \cmidrule(lr){8-9}
        & SPL & SSPL & SPL & SSPL & SPL & SSPL & SPL & SSPL \\
      \midrule
      RoboHop~\cite{RoboHop}
        & 36.34 & 54.25
        & 54.51 & 69.98
        & 60.57 & 68.29
        & 57.52 & 68.47 \\
      \oursMSF{} (Ours)
        & \cellcolor{blue!15}\textbf{63.60} & \cellcolor{blue!15}\textbf{78.84}
        & \cellcolor{blue!15}\textbf{63.60} & \cellcolor{blue!15}\textbf{78.33}
        & \cellcolor{blue!15}\textbf{66.62} & \cellcolor{blue!15}\textbf{75.20}
        & \cellcolor{blue!15}\textbf{63.56} & \cellcolor{blue!15}\textbf{73.89} \\
      \bottomrule
    \end{tabular}%
  }
  \vspace{0.5em}
  \caption{Navigation performance comparison.  The best value in each column is highlighted in blue.}
  \label{tab:navigation}
\end{table}

\paragraph{Impact of the Feature Encoder}
To isolate the contribution of the feature encoder in our pipeline, we replace \master{}’s cross–attention block with two alternative, purely 2D backbones: \textit{CroCo} (shared by both \master{} and \oursMSF{}) and \textit{DINOv2}, a state-of-the-art encoder often used for off-the-shelf segment matching~\cite{conceptgraph,RoboHop,garg2024revisitanything}. 
As reported in \Cref{tab:segmatch_ablation_scannetpp}, neither backbone yields competitive segment-matching accuracy.  
This result underscores that learning segment descriptors from a 2D foundation model alone is inadequate; geometric context is essential.  
In contrast, \master{}’s cross–attention layers, 3D-aware training regimen, and explicit formulation of image matching in 3D jointly endow the model with the priors required for reliable instance association. The fact that CroCo is always second best suggests that cross-view-completion, by itself, can yield superior results for segment matching between image pairs.

\begin{table*}[ht!]
\centering
\small
\resizebox{\textwidth}{!}{%
\begin{tabular}{>{\raggedright\arraybackslash}p{3cm}
                  ccc   %
                  ccc   %
                  ccc   %
                  ccc}  %
\toprule
\textbf{Method}
 & \multicolumn{3}{c}{\textbf{0\textdegree–45\textdegree}}
 & \multicolumn{3}{c}{\textbf{45\textdegree–90\textdegree}}
 & \multicolumn{3}{c}{\textbf{90\textdegree–135\textdegree}}
 & \multicolumn{3}{c}{\textbf{135\textdegree–180\textdegree}} \\
\cmidrule(lr){2-4} \cmidrule(lr){5-7} \cmidrule(lr){8-10} \cmidrule(lr){11-13}
 & AUPRC & R@1 & R@5
 & AUPRC & R@1 & R@5
 & AUPRC & R@1 & R@5
 & AUPRC & R@1 & R@5 \\
\midrule
DINOv2-SegFeat
 & 64.7 & 71.5 & 89.3
 & 55.7 & 65.9 & 87.3
 & 45.4 & 59.1 & 84.4
 & 36.8 & 53.4 & 81.3 \\

CroCo-SegFeat
 & \second{73.4} & \second{78.8} & \second{92.3}
 & \second{64.0} & \second{73.1} & \second{90.6}
 & \second{50.7} & \second{64.6} & \second{87.5}
 & \second{38.5} & \second{56.6} & \second{84.1} \\

\oursMSF{} (Ours)
 & \best{92.8} & \best{93.6} & \best{98.0}
 & \best{91.1} & \best{92.2} & \best{97.6}
 & \best{88.0} & \best{89.5} & \best{96.8}
 & \best{83.6} & \best{85.9} & \best{95.9} \\
\bottomrule
\end{tabular}%
} %
\caption{Proposed method and model ablations performance across pose‐bins on ScanNet++~\cite{yeshwanthliu2023scannetpp}. \colorbox{blue!20}{Blue} cells mark the best scores; \colorbox{orange!20}{Orange} cells mark the second‐best.}
\label{tab:segmatch_ablation_scannetpp}
\end{table*}

\paragraph{Qualitative Results}

\begin{figure}
  \centering
  \begin{tabular}{ccc ccc}
  \multicolumn{3}{c}{\includegraphics[width=0.48\textwidth]{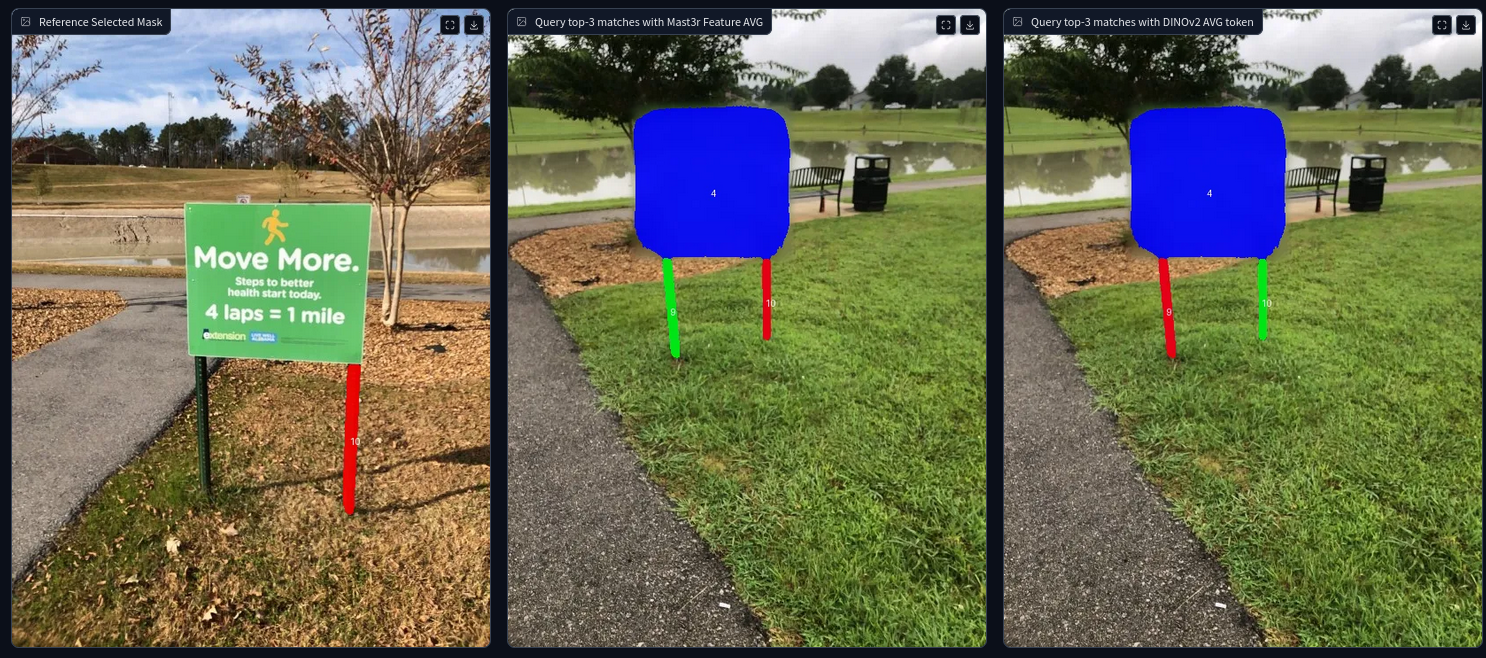}} &  \multicolumn{3}{c}{\includegraphics[width=0.48\textwidth]{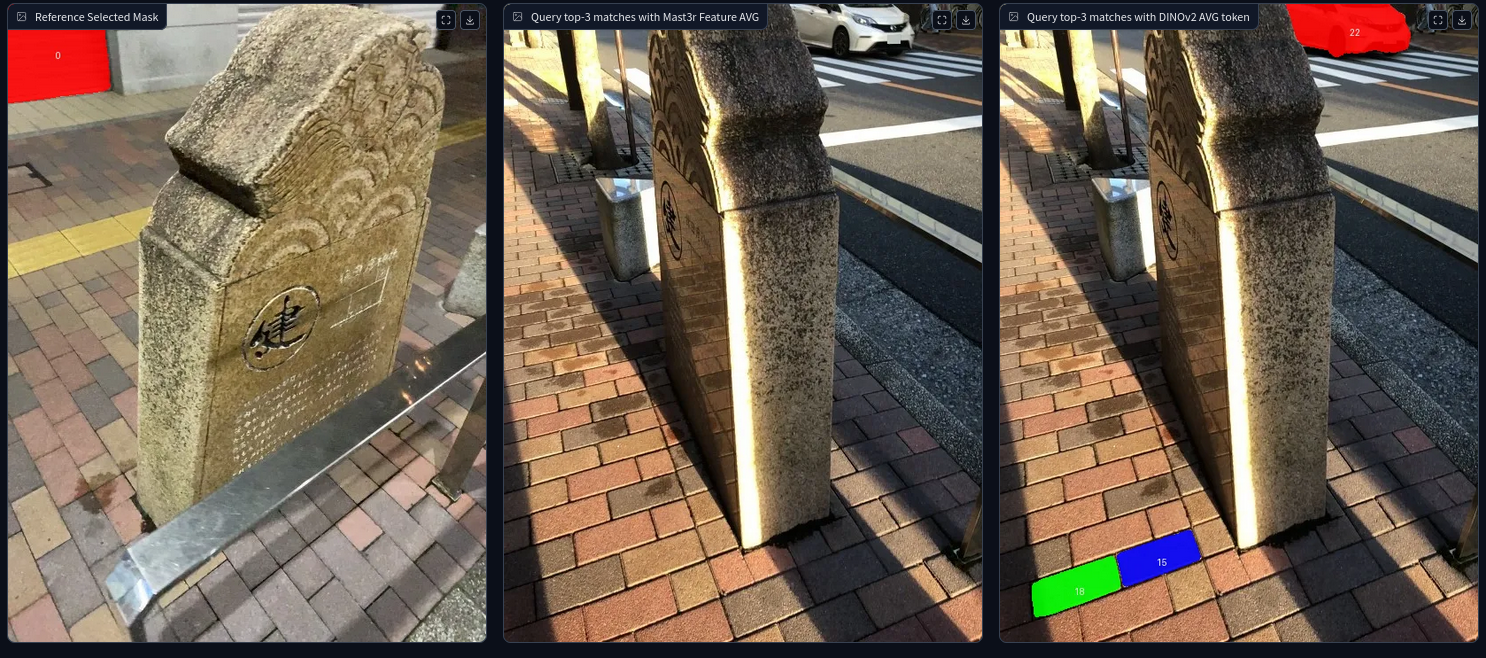}}
 \\
  \hspace{0.6cm} Query & \hspace{0.4cm} SegMASt3R & DINOv2 & 
  \hspace{0.6cm} Query & \hspace{0.4cm} SegMASt3R & DINOv2 \\
  \end{tabular}
  \caption{\textit{MapFree Outdoor Dataset} - \textbf{Perceptual Instance Aliasing} (left): the right leg of the signboard as a query segment (red) is correctly matched by our method but mismatched with its left leg by DINOv2. \textbf{Sinkhorn Matches to Dustbin} (right): the query segment (red) is not visible in the reference image and is correctly ignored by our method, whereas DINOv2 mismatches it with a vehicle segment.}
  \label{fig:qual-segfeat-mapfree}
\end{figure}

\begin{figure}%
    \centering
    \includegraphics[width=0.9\textwidth]{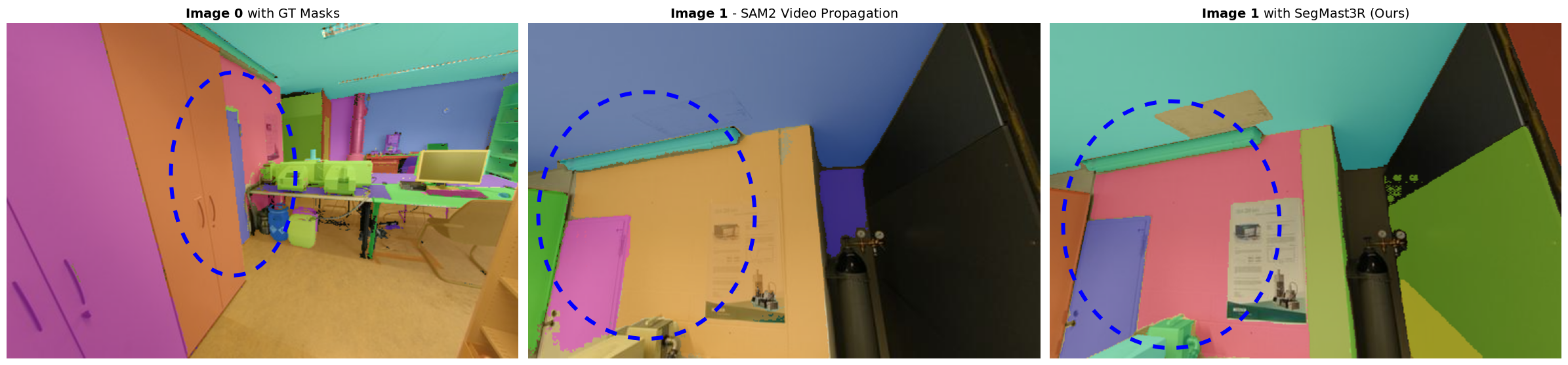}\vspace{1mm}
    \includegraphics[width=0.9\textwidth]{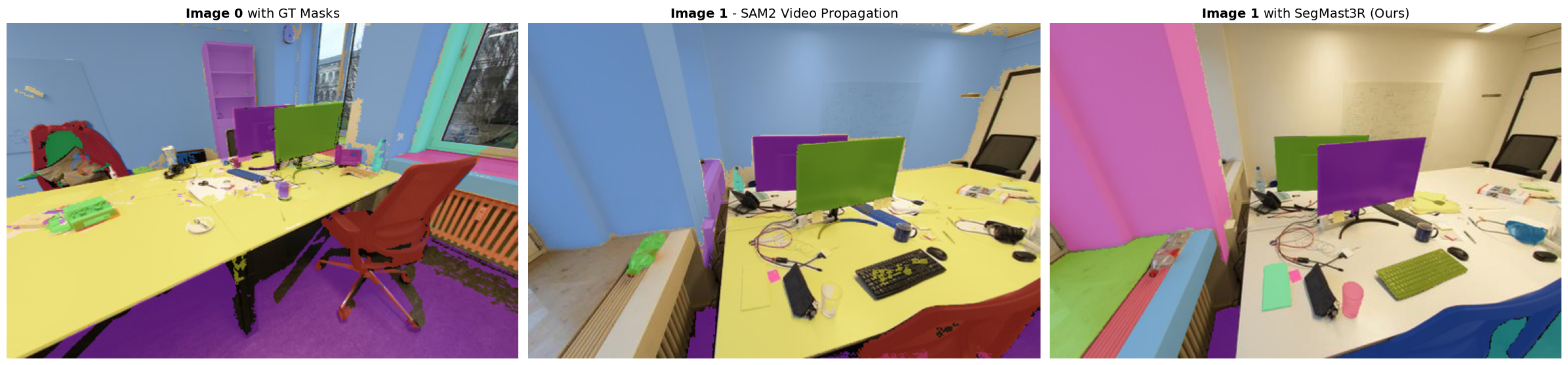}
    \caption{\textit{ScanNet++ Dataset} -- \textbf{Wide-baseline Matching} (top): The wall (pink) and the door (blue) in the query image (left) gets incorrectly associated by SAM2's video propagation (middle), whereas SegMASt3R (right) is able to correctly match them despite very limited visual overlap. \textbf{Perceptual Instance Aliasing} (bottom): unlike SAM2, SegMASt3R is able to correctly associate the pair of monitors under the simultaneous duress of an opposing viewpoint observation and perceptual instance aliasing.}
    \label{fig:spp-qual}
\end{figure}

Figure~\ref{fig:spp-qual} compares \oursMSF{} with SAM2's two-frame video propagation-based segment matching under \textit{extreme viewpoint variations} on the ScanNet++ dataset. Each row shows a reference image, SAM2's matches, \oursMSF{}'s matches, respectively. In the top row, a wall (pink) and a door (blue) are mismatched by SAM2, whereas \oursMSF{} correctly associates them, despite a very limited visual overlap. In the bottom row, SAM2 gets confused between the two monitors, whereas \oursMSF{} is able to correctly associate them despite the simultaneous effect of $180^\circ$ viewpoint shift and \textit{perceptual instance aliasing} (i.e., different instances of the same object category in an image potentially lead to mismatches).  
Figure~\ref{fig:qual-segfeat-mapfree} presents qualitative results on the (outdoor) MapFree~\cite{arnold2022mapfree} dataset, where we compare our ScanNet-trained \oursMSF{} with the off-the-shelf DINOv2~\cite{oquab2023dinov2} features. Each image triplet represents the query segment (left) and its retrieved matches using our method (middle) and DINOv2 (right). The query segment is displayed in red color, and its top three matches are respectively displayed in red, green, and blue colors. The left image triplet illustrates another case of perceptual instance aliasing. \oursMSF{} is able to resolve this aliasing problem, whereas DINOv2 confuses the right and left legs of the signboard. The right image panel shows that our Sinkhorn solver effectively learns dustbin allocations for explicitly rejecting negatives, that is, the segments which do not have any corresponding match, whereas DINOv2 leads to incorrect matches.

\section{Conclusion}
We proposed~\oursMSF{}, a simple method to re-purpose an existing 3D foundation model MASt3R for image segment matching. Our proposed method achieves excellent results on ScanNet++ and Replica with a simple pipeline and a minimal amount of training. It especially excels on wide-baseline segment matching between image pairs. In addition, we show that~\oursMSF{} has practical applicability by evaluating it's performance on the downstream tasks of 3D instance mapping and object-relative topological navigation, where we significantly outperform the corresponding baselines. Overall, this work attempts to establish segment matching as a core computer vision capability, which will enable even more downstream applications in future alongside the advances in image segmentation and data association.  
\newpage

\bibliographystyle{plainnat}
\bibliography{neurips_2025}

\clearpage  %
\appendix

\part*{Supplementary Material}

\newcommand{\obs}{I}           %
\newcommand{\orgb}{\obs^{\text{rgb}}}  %
\newcommand{\odep}{\obs^{\text{depth}}} %
\newcommand{\opos}{\theta}  %
\newcommand{\omask}{\mathbf{m}} %
\newcommand{\mfeat}{\mathbf{f}} %
\newcommand{\objf}{\mathbf{f_o}} %
\newcommand{\mbsp}{\mathbf{p}}  %
\newcommand{\objp}{\mathbf{p_o}} %

\newcommand{\ibox}{\mathbf{b}} %
\newcommand{\pbox}{\ibox^{\text{3D}}} %

\newcommand*{\seg}{\ensuremath{\text{Seg}}} %
\newcommand{\SE}{\text{Embed}}        %
\newcommand{\lang}{\text{LVLM}}        %
\newcommand{\glang}{\text{LLM}}        %

\newcommand{\objmap}{\mathcal{M}} %
\newcommand{\obj}{\mathbf{o}}      %
\newcommand{\objset}{\mathbf{O}}   %
\newcommand{\edg}{\mathbf{e}}      %
\newcommand{\edgeset}{\mathbf{E}}  %
\newcommand{\capp}{\mathbf{c}}      %
\newcommand{\rcapp}{\hat{\capp}}      %

\newcommand{\semSim}{\phi_{\text{sem}}}  %
\newcommand{\geoSim}{\phi_{\text{geo}}}  %
\newcommand{\nnratio}{\text{nnratio}}  %
\newcommand{\overallSim}{\phi}          %
\newcommand{\simMat}{\mathbf{\phi}}     %
\newcommand{\newofeat}{\objf_{\text{new}}}  %
\newcommand{\bbodt}{\epsilon_b} %
\newcommand{\bbmat}{\mathbf{\psi}} %
\newcommand{\oldofeat}{\objf_{\text{old}}}  %
\newcommand{\numDet}{n}              %
\newcommand{\sminus}{\text{-}}  %

\newcommand{\imgseq}{\mathcal{I}} %

\newcommand{\wedgev}{w}      %
\newcommand{\ovThresh}{\alpha}  %

\newcommand{\mstree}{\mathcal{T}}  %

\section{Cross-Dataset Generalization on Replica}
To evaluate cross-dataset generalization, we train our model on ScanNet++\cite{yeshwanthliu2023scannetpp} and directly evaluate it on the Replica dataset\cite{replica19arxiv}, without any fine-tuning. Table~\ref{tab:segmatch_posebins_replica_addn} reports a detailed comparison against competitive baselines, highlighting the robustness of our approach under distribution shift.

\begin{table*}[htbp]
\centering
\small
\resizebox{\textwidth}{!}{%
\begin{tabular}{%
  >{\raggedright\arraybackslash}p{2cm}  %
  >{\raggedright\arraybackslash}p{3cm}  %
  ccc   %
  ccc   %
  ccc   %
  ccc}  %
\toprule
\textbf{Type}
 & \textbf{Method}
 & \multicolumn{3}{c}{\textbf{0\textdegree–45\textdegree}}
 & \multicolumn{3}{c}{\textbf{45\textdegree–90\textdegree}}
 & \multicolumn{3}{c}{\textbf{90\textdegree–135\textdegree}}
 & \multicolumn{3}{c}{\textbf{135\textdegree–180\textdegree}} \\
\cmidrule(lr){3-5} \cmidrule(lr){6-8} \cmidrule(lr){9-11} \cmidrule(lr){12-14}
 &  & AUPRC & R@1 & R@5
      & AUPRC & R@1 & R@5
      & AUPRC & R@1 & R@5
      & AUPRC & R@1 & R@5 \\
\midrule
\multirow{4}{2cm}{Local Feature\\Matching (LFM)}
 & SP-LG~\cite{detone18superpoint,lindenberger2023lightglue}
   & 64.69 & 67.89 & 72.3
   & 46.89 & 51.59 & 56.1
   & 21.86 & 28.74 & 32.83
   & 21.82 & 28.72  & 30.9 \\

 & GiM-DKM~\cite{xuelun2024gim,edstedt2023dkm}
   & 75.46 & 81.38 & 83.51
   & 67.26 & 74.08 & 76.8
   & 43.94 & 55.8 & 61.41
   & 35.37 & 49.41 & 53.86 \\

 & RoMA~\cite{edstedt2024roma}
   & 77.3 & 84.06 & 86.86
   & 69.35 & \second{79.08} & 84
   & 46.69 & \second{68.49} & 77.48
   & 31.79 & 60.48 & 68.94 \\

 & \master~\cite{leroy2024grounding}
   & 78.2 & \second{86.5} & 89.4
   & \second{69.5} & 77.6 & 81.0
   & \second{48.0} & 60.4 & 64.6
   & 32.5 & 49.0 & 54.1 \\
\midrule
\multirow{4}{2cm}{Segment Matching\\(SegMatch)}
 & SAM2~\cite{ravi2024sam}
   & \second{80.09} & 82.28 & 84.61
   & 54.58 & 62.03 & 65.41
   & 40.69 & 53.72 & 56.67
   & \second{37.78} & 54.59 & 56.42 \\

 & DINOv2~\cite{oquab2023dinov2}
   & 55.85 & 74.25 & \second{96.55}
   & 31.12 & 59.64 & \second{92.84}
   & 21.71 & 57.68 & \second{92.33}
   & 17.29 & 59.28 & \second{89.64} \\

 & SegVLAD~\cite{garg2024revisitanything}
   &  67.49 & 77.65 & 95.52
   & 46.84 & 66.46 & 92.26
   & 40.21 & 67.23 & 91.81
   & 37.8 & \second{69.6} & 89.15 \\

 & \master~\cite{leroy2024grounding}
   & 52.2 & 57.5 & 81.2
   & 39.1 & 51.0 & 78.6
   & 23.6 & 45.9 & 77.2
   & 17.2 & 43.8 & 75.7 \\
\midrule
Ours
 & \oursMSF{}
   & \best{95.0} & \best{96.0} & \best{98.6}
   & \best{86.2} & \best{91.2} & \best{96.4}
   & \best{73.4} & \best{85.2} & \best{95.7}
   & \best{68.4} & \best{83.8} & \best{94.8} \\

\bottomrule
\end{tabular}%
} %
\caption{Performance of additional baselines across pose‐bins on Replica\cite{replica19arxiv}. \colorbox{blue!20}{Blue} cells mark the best scores; \colorbox{orange!20}{Orange} cells mark the second‐best.}
\label{tab:segmatch_posebins_replica_addn}
\end{table*}

\section{Local-Feature Matching (LFM) Baseline Setup}
\begin{algorithm}[htbp]
\caption{Segment Correspondence via Keypoint Voting}
\label{alg:segment-matching}
\begin{algorithmic}[1]
\Require Images $I_0, I_1$; segment masks $\mathcal{M}_0 \in \{0,1\}^{M \times H \times W}$, $\mathcal{M}_1 \in \{0,1\}^{N \times H \times W}$; keypoint matcher $\mathcal{A}$; max matches $K$
\Ensure Segment assignment vector $\hat{\mathcal{C}} \in \mathbb{N}^M$ (with \texttt{-1} indicating no confident match)
\State $\{(x_i^0, y_i^0), (x_i^1, y_i^1)\}_{i=1}^K \gets \mathcal{A}(I_0, I_1)$
\State Initialize vote matrix $V \in \mathbb{Z}^{M \times N} \gets 0$
\For{$i = 1$ to $K$}
    \State Find $m_i$ such that $\mathcal{M}_0[m_i, y_i^0, x_i^0] = 1$
    \State Find $n_i$ such that $\mathcal{M}_1[n_i, y_i^1, x_i^1] = 1$
    \If{$m_i$ and $n_i$ are valid}
        \State $V[m_i, n_i] \gets V[m_i, n_i] + 1$
    \EndIf
\EndFor
\For{$m = 1$ to $M$}
    \If{$\sum_n V[m, n] = 0$}
        \State $\hat{\mathcal{C}}[m] \gets -1$ \Comment{No confident match}
    \Else
        \State $\hat{\mathcal{C}}[m] \gets \arg\max_n V[m, n]$
    \EndIf
\EndFor
\State \Return $\hat{\mathcal{C}}$
\end{algorithmic}
\end{algorithm}

A range of local feature matchers (LFMs) exist at varying densities—SuperPoint~\cite{detone18superpoint} (sparse), LoFTR~\cite{sun2021loftr} (semi-dense), and RoMa or MASt3R~\cite{edstedt2024roma,leroy2024grounding} (dense). We leverage the EarthMatch (IMM)~\cite{Berton_2024_EarthMatch} toolkit, which provides a unified and modular interface for applying a wide range of local feature matchers. This abstraction significantly simplifies the integration of keypoint-based methods into our pipeline, allowing us to systematically evaluate segment correspondence performance across multiple matcher densities.

We use a \textit{voting-based aggregation} scheme (Algorithm~\ref{alg:segment-matching}) to derive segment-level correspondences via LFMs. Given two images and their respective binary segment masks $\mathcal{M}_0 \in \{0,1\}^{M \times H \times W}$ and $\mathcal{M}_1 \in \{0,1\}^{N \times H \times W}$, a chosen LFM $\mathcal{A}$ produces $K$ matched keypoint pairs. Each matched pair votes for the segment it belongs to in the source and target masks, incrementing a \textit{vote matrix} $V \in \mathbb{Z}^{M \times N}$.

Finally, segment-level correspondences are assigned by selecting, for each source segment $m$, the target segment $n$ with the highest vote count. Segments with no valid votes are marked as unmatched (assigned label \texttt{-1}). This voting mechanism provides a simple yet effective bridge between pixel-level keypoint matches and object-level segment associations.

\section{Additional Qualitative Results on ScanNet++ Dataset}
\begin{figure}[htbp]
    \centering

    \begin{subfigure}[t]{0.48\textwidth}
        \centering
        \includegraphics[width=\linewidth]{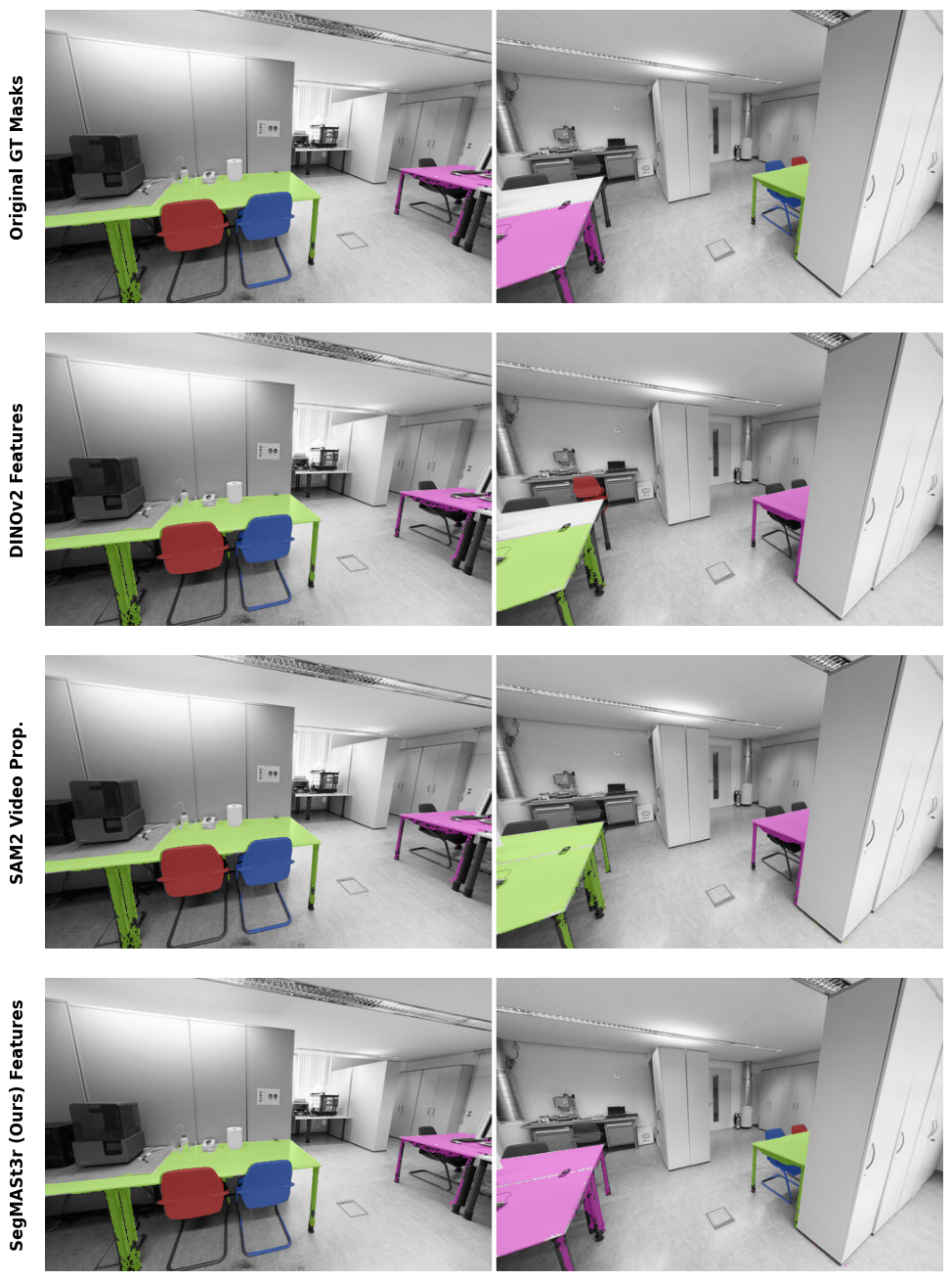}
        \caption{}
        \label{fig:spp-1}
    \end{subfigure}
    \hfill
    \begin{subfigure}[t]{0.48\textwidth}
        \centering
        \includegraphics[width=\linewidth]{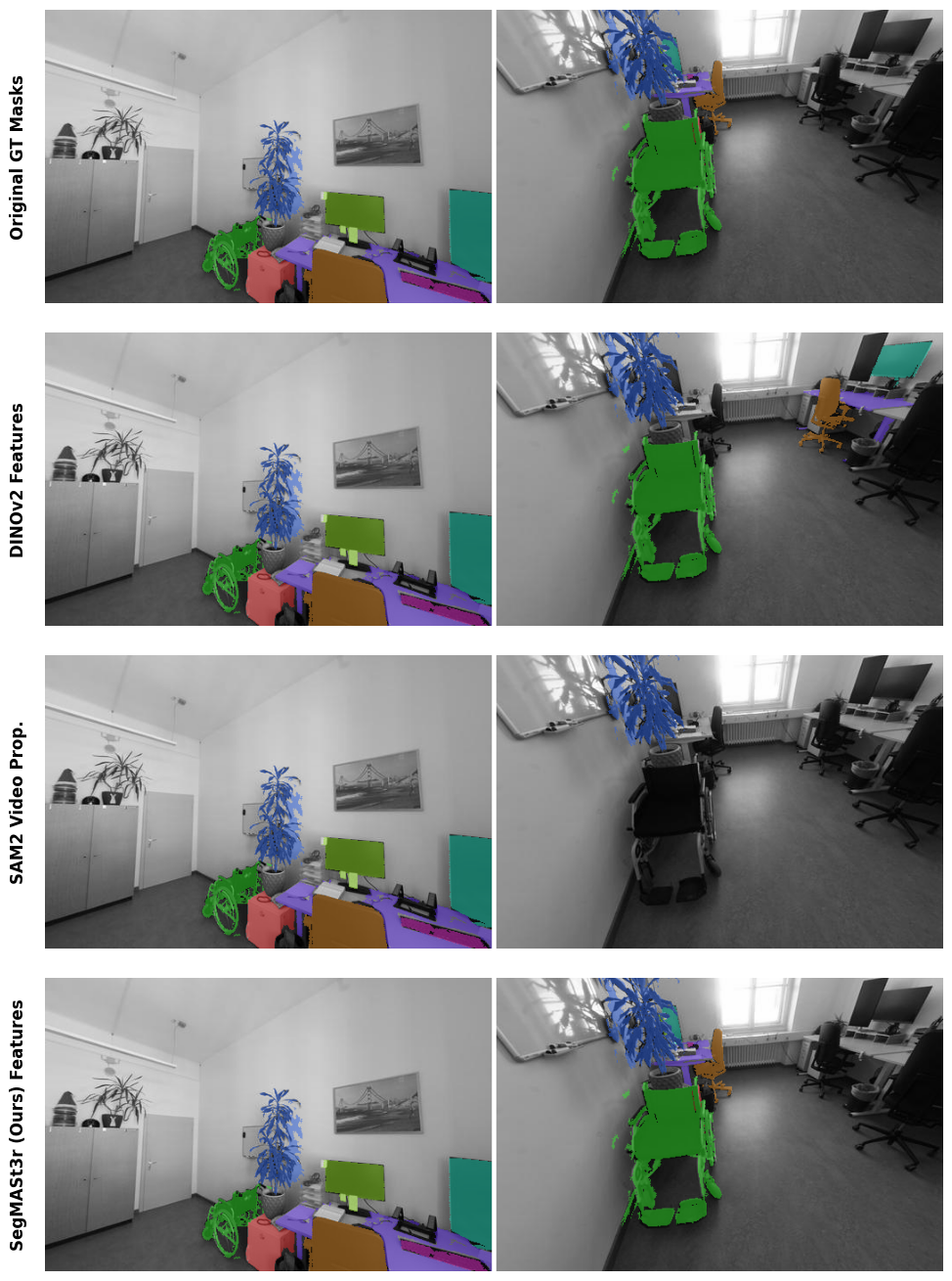}
        \caption{}
        \label{fig:spp-2}
    \end{subfigure}

    \begin{subfigure}[t]{0.48\textwidth}
        \centering
        \includegraphics[width=\linewidth]{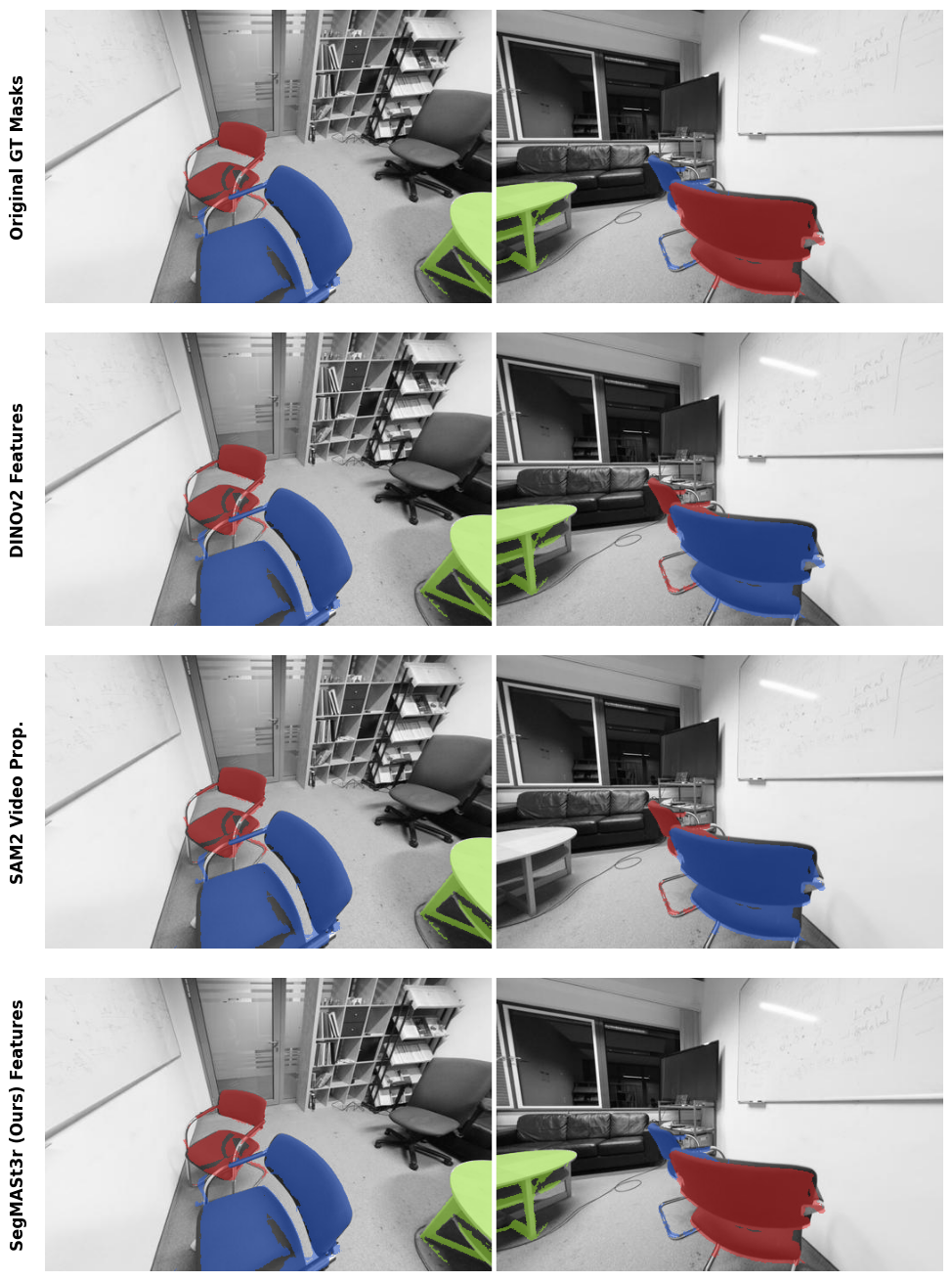}
        \caption{}
        \label{fig:spp-9}
    \end{subfigure}
    \hfill
    \begin{subfigure}[t]{0.48\textwidth}
        \centering
        \includegraphics[width=\linewidth]{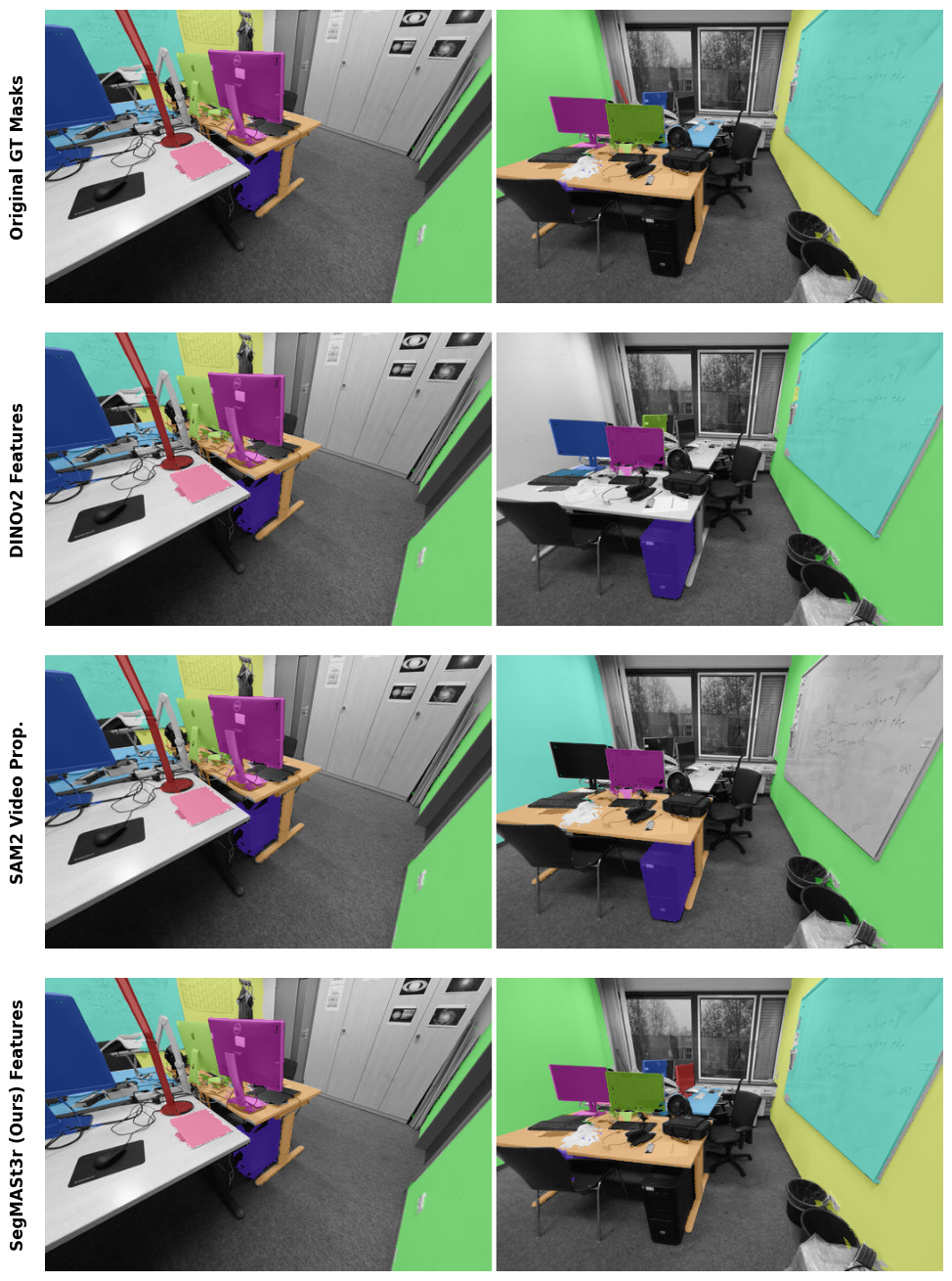}
        \caption{}
        \label{fig:spp-8}
    \end{subfigure}

    \caption{More examples comparing proposed method against DINOv2~\cite{oquab2023dinov2} and SAM2~\cite{ravi2024sam} for segment-matching on the ScanNet++ Dataset~\cite{yeshwanthliu2023scannetpp} under \textbf{Wide-baseline conditions [135--180\textdegree{} viewpoint change]}. Both baselines tend to incorrectly assign segment correspondences given an opposite viewing direction-the tables (pink and green) in (a) as well as the chairs (red and blue) in (c). Another failure mode SAM2 specifically exhibits is the inability to propagate masks in challenging view-point change settings as seen for the wheelchair (green) in (b). In contrast, the proposed method demonstrates accurate segment matching, attributed to its conditioning on 3D-aware priors. \\ \textit{Note: Images have been greyed out to improve the visibility of the segments in question.}}
    \label{fig:spp-qual-supp}
\end{figure}

We present additional qualitative results (Figure~\ref{fig:spp-qual-supp}) on the ScanNet++ dataset\cite{yeshwanthliu2023scannetpp}, highlighting our method’s robustness to perceptual instance aliasing and wide-baseline viewpoint changes. These examples further demonstrate accurate segment correspondences despite challenging geometric and appearance ambiguities.

\section{Details on Instance-Mapping}

\begin{table}[htbp]
  \centering
  \setlength{\tabcolsep}{4pt}
  \renewcommand{\arraystretch}{1.2}
  \resizebox{\linewidth}{!}{%
\begin{tabular}{l*{8}{c}}
  \toprule
  \textbf{Method}
    & \textbf{office0}
    & \textbf{office1}
    & \textbf{office2}
    & \textbf{office3}
    & \textbf{office4}
    & \textbf{room0}
    & \textbf{room1}
    & \textbf{room2} \\
  \cmidrule(lr){2-2}
  \cmidrule(lr){3-3}
  \cmidrule(lr){4-4}
  \cmidrule(lr){5-5}
  \cmidrule(lr){6-6}
  \cmidrule(lr){7-7}
  \cmidrule(lr){8-8}
  \cmidrule(lr){9-9}
  
    & {AP / AP@50}
    & {AP / AP@50}
    & {AP / AP@50}
    & {AP / AP@50}
    & {AP / AP@50}
    & {AP / AP@50}
    & {AP / AP@50}
    & {AP / AP@50} \\
  \midrule
  ConceptGraphs (MobileSAM Masks)~\cite{conceptgraph}
    & 11.84 / 28.43
    & 20.31 / 43.79
    & 8.63 / 22.82
    & 8.07 / 22.83
    & 9.46 / 24.73
    & 12.23 / 34.34
    & 5.83 / 12.96
    & 7.83 / 23.82 \\
  ConceptGraphs (GT Masks)~\cite{conceptgraph}
    & 43.53 / 69.68
    & 22.48 / 40.71
    & 43.46 / 60.69
    & 32.06 / 53.44
    & 39.63 / 68.22
    & 44.89 / 69.64
    & 17.96 / 36.53
    & 25.93 / 43.63 \\
  \oursMSF (Ours, GT Masks)
    & \cellcolor{blue!15}79.93 / \cellcolor{blue!15}87.17
    & \cellcolor{blue!15}54.89 / \cellcolor{blue!15}64.42
    & \cellcolor{blue!15}64.00 / \cellcolor{blue!15}85.50
    & \cellcolor{blue!15}58.02 / \cellcolor{blue!15}79.93
    & \cellcolor{blue!15}67.48 / \cellcolor{blue!15}85.01
    & \cellcolor{blue!15}71.02 / \cellcolor{blue!15}91.22
    & \cellcolor{blue!15}64.09 / \cellcolor{blue!15}85.50
    & \cellcolor{blue!15}56.35 / \cellcolor{blue!15}76.66 \\
  \bottomrule
\end{tabular}
}
  \vspace{0.5em}
  \caption{Class-Agnostic instance-mapping performance (AP and AP@50) on Replica scenes, shown in percentage.  The best value in each column is highlighted in blue.}
  \label{tab:instance_mapping_replica_supplementary}
\end{table}

By default, ConceptGraphs~\cite{conceptgraph} is evaluated on the scanned RGB-D trajectories of the Replica~\cite{replica19arxiv} as provided by Nice-SLAM~\cite{Zhu2022CVPR}. However, this version does not include ground-truth instance masks and instead relies on MobileSAM~\cite{mobile_sam} for mask generation. We first evaluate the method using this setup, which corresponds to the first row of Table~\ref{tab:instance_mapping_replica_supplementary}. To assess performance with ground-truth instance segmentation, we use the pre-rendered Replica RGB-D sequence released by Semantic-NeRF~\cite{Zhi:etal:ICCV2021}, which provides 900 RGB-D frames per scene along with ground-truth instance masks and camera poses. We subsample each sequence with a stride of 5 and run ConceptGraphs on a total of 180 frames per scene. This evaluation setting corresponds to the second row of Table~\ref{tab:instance_mapping_replica_supplementary}. We observe that the trend remains consistent, with \oursMSF{} continuing to outperform ConceptGraphs on the instance mapping task.

\subsection{Evaluation Methodology for Instance Mapping}

We assess the accuracy of instance mapping by how well individual object geometries are preserved, \ie \textit{quality of the instance boundaries and whether they are multi-view consistent}. We follow the class-agnostic evaluation protocol as done by OpenYOLO3D \cite{boudjoghra2025openyolo}, which benchmarks generic 3D object instances in the given scene. Formally, given a predicted 3D instance-map point-cloud and the corresponding ground-truth, we compute the Average Precision (AP) over a range of Intersection-over-Union (IoU) thresholds.  In particular, we report:
\begin{itemize}
  \item \textbf{AP@0.50} reports precision when IoU\,=\,0.50, measuring correctness at a moderate overlap.
  \item \textbf{AP} is the mean precision over IoU thresholds from 0.50 to 0.95 in steps of 0.05, capturing both coarse and fine alignment.
\end{itemize}

\begin{figure}[ht]
    \centering
    \begin{subfigure}[b]{0.28\linewidth}
        \centering
        \includegraphics[height=2.8cm, keepaspectratio]{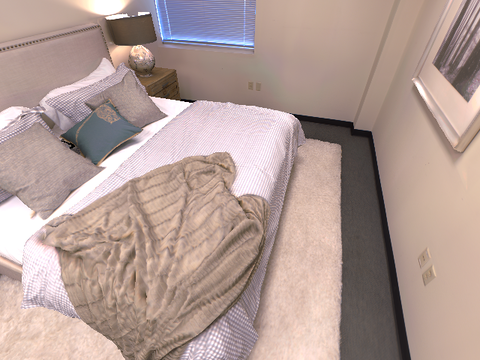}
        \caption{Ground Truth}
    \end{subfigure}
    \hfill
    \begin{subfigure}[b]{0.28\linewidth}
        \centering
        \includegraphics[height=2.8cm, keepaspectratio]{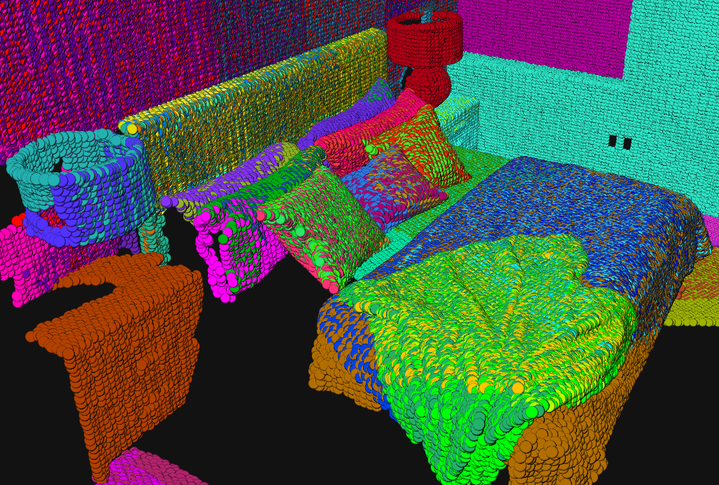}
        \caption{ConceptGraphs}
    \end{subfigure}
    \hfill
    \begin{subfigure}[b]{0.28\linewidth}
        \centering
        \includegraphics[height=2.8cm, keepaspectratio]{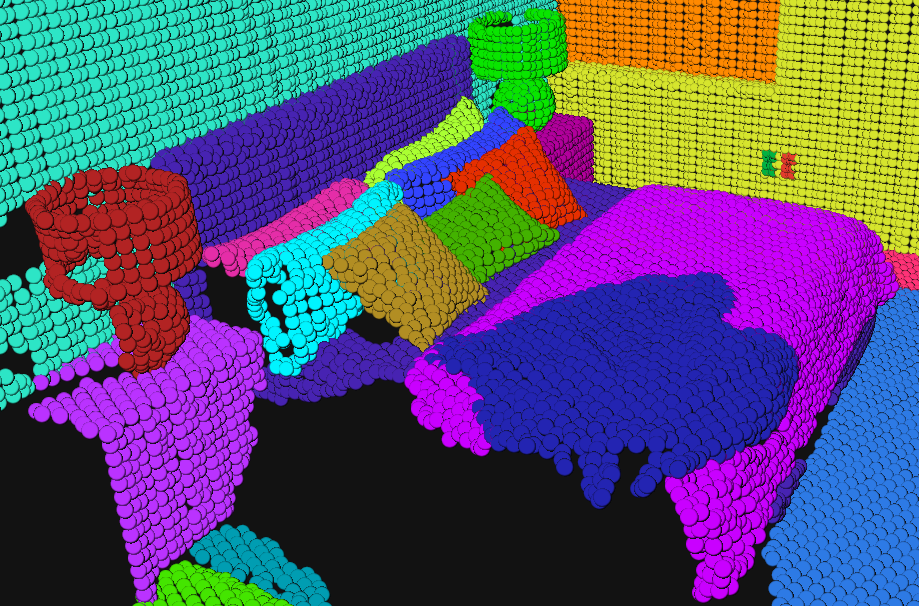}
        \caption{Ours}
    \end{subfigure}
    \caption{\textbf{Qualitative comparison of instance maps.} Each color corresponds to a different object instance. The ground truth (left) provides the closest reference RGB view of the scene from the RGB sequence. ConceptGraphs (middle) tends to over-segment objects, assigning multiple instance IDs to the same object, resulting in fragmented coloring (\textit{e.g. brown duvet covered by green and yellow labels}). Our method (right) produces cleaner and more consistent instance groupings.}
    \label{fig:instance_map_comparison_full}
\end{figure}

\subsection{ConceptGraphs for Instance Mapping}
We use Concept-Nodes, a lightweight implementation of ConceptGraphs~\cite{conceptgraph}. %
We generate an instance-level 3D map by aggregating detections over time and associating them with previously observed objects in a global map via semantic and geometric matching.

\subsubsection{Object-Centric 3-D Representation}
Given a stream of RGB-D observations
$\mathcal{I}=\{I_1,\dots,I_T\}$,
ConceptGraphs incrementally builds a 3-D scene graph
$\mathcal{M}_t=(\mathcal{O}_t)$.
Each node $o_j\in\mathcal{O}_t$ is represented by a point cloud
$\mathbf{P}_{o_j}$ and a unit-normalised semantic descriptor
$\mathbf{f}_{o_j}$.
At time $t$ the incoming frame
$I_t=\langle I_t^{\mathrm{rgb}}, I_t^{\mathrm{depth}},\boldsymbol{\theta}_t\rangle$
(colour, depth, pose) is fused into the map by either updating an existing object or instantiating a new one.

\subsubsection{Class-Agnostic 2-D Segmentation}
For each frame we extract $M$ class-agnostic masks
$\{\mathcal{M}_{t,i}\}_{i=1}^{M}
      =\mathrm{Seg}(I_t^{\mathrm{rgb}})$
using \textsc{YOLO-World}~\cite{Cheng2024YOLOWorld} and \textsc{MobileSAM}~\cite{mobile_sam} (or ground truth, when available). Every mask is fed to a CLIP~\cite{radford2021learning} encoder to obtain a visual descriptor
$\mathbf{f}_{t,i}=\mathrm{SE}(I_t^{\mathrm{rgb}},\mathcal{M}_{t,i})$.
The masked RGB-D region is back-projected and transformed to the
global frame, yielding a point cloud $\mathbf{P}_{t,i}$ paired with
$\mathbf{f}_{t,i}$.

\subsubsection{Object Association}
For every newly detected object candidate 
$\langle \mathbf{P}_{t,i},\mathbf{f}_{t,i}\rangle$ 
we measure its similarity to each map object
$\langle \mathbf{P}_{j},\mathbf{f}_{j}\rangle$ that exhibits spatial overlap.  
Geometric consistency is captured by the \emph{nearest-neighbour ratio}
\[
s_{\text{geo}}(i,j)=
\operatorname{NNRatio}(\mathbf{P}_{t,i},\mathbf{P}_{j}) ,
\]
defined as the fraction of points in the candidate cloud whose closest
point in the reference cloud lies within the Euclidean tolerance
$\delta_{\text{nn}}$.  
Appearance similarity is obtained from the cosine similarity of the
CLIP descriptors, rescaled from $[-1,1]$ to $[0,1]$,
\[
s_{\text{sem}}(i,j)=\tfrac12\bigl(\mathbf{f}_{t,i}^{\!\top}\mathbf{f}_{j}\bigr)+\tfrac12 .
\]
The two scores are linearly blended into a fused similarity
\[
s(i,j)=\alpha\,s_{\text{sem}}(i,j)+(1-\alpha)\,s_{\text{geo}}(i,j),
\qquad \alpha\!\in\![0.1,0.5] ,
\]
where the weight $\alpha$ balances semantic and geometric evidence and
is tuned per scene.  
Greedy matching assigns each detection to the map object with the
highest fused similarity; if this peak score falls below the threshold
$\delta_{\text{sim}}\in[0.90,0.96]$, the detection is instead used to
instantiate a new object instance.  In practice, we sweep and select $\delta_{\text{sim}}$ in the range
$0.90\dots0.96$ to accommodate scene-specific variability in texture
and clutter.

\subsubsection{Object Fusion}
Once a detection $i$ has been matched to map object $o_j$,
its semantic and geometric information is merged into the map.
Let $n_j$ denote the number of detections previously fused into
$o_j$ and $\delta_{\text{voxel}}$ the voxel size used for
down-sampling.

\paragraph{Semantic update.}
We maintain a running average of CLIP descriptors so that each new
observation contributes equally:
\[
\mathbf{f}_{j}\;\leftarrow\;
\frac{n_{j}\,\mathbf{f}_{j}+\mathbf{f}_{t,i}}{n_{j}+1},
\qquad n_{j}\;\leftarrow\; n_{j}+1.
\]

\paragraph{Geometric update.}
The point cloud of the object is augmented and then compacted to
prevent redundancy:
\[
\mathbf{P}_{j}\;\leftarrow\;
\operatorname{Downsample}\!\bigl(
      \mathbf{P}_{j}\;\cup\;\mathbf{P}_{t,i},
      \delta_{\text{voxel}}
\bigr),
\]
where $\operatorname{Downsample}(\cdot,\delta_{\text{voxel}})$ groups
points whose pairwise distances are below the voxel size
($\delta_{\text{voxel}}\!=\!1\text{\,cm}$ in all experiments) and
replaces each group by its centroid.

\subsubsection{\oursMSF{} for Instance Mapping}
We use \textsc{\oursMSF{}} to directly establish object-level correspondences across image pairs, based on both mask-level appearance features and geometric consistency. Unlike Concept-Graphs, which incrementally associates detections with a growing map, our method operates pairwise and enforces geometric alignment through 3D checks.

\paragraph{Pairwise Mask Matching.}  
Given two RGB-D frames $I_s = \langle I_s^{\text{rgb}}, I_s^{\text{depth}}, \theta_s \rangle$ and $I_t = \langle I_t^{\text{rgb}}, I_t^{\text{depth}}, \theta_t \rangle$ and their respective sets of ground-truth instance masks $\{\omask_{s,i}\}_{i=1}^{M_s}$ and $\{\omask_{t,j}\}_{j=1}^{M_t}$, we extract visual descriptors for each mask using our method, \oursMSF{}:
\[
\mfeat_{s,i}, \mfeat_{t,j} = \textsc{\oursMSF{}}(I_s^{\text{rgb}},I_t^{\text{rgb}}, \omask_{s,i}, \omask_{t,j})
\]
where $\mfeat_{s,i}, \mfeat_{t,j} \in \mathbb{R}^{24}$ are the mask feature vectors.
These features are used to construct a pairwise similarity matrix, and a soft assignment between masks is computed using the Sinkhorn algorithm. Tentative correspondences are obtained by row-wise $\arg\max$ over the resulting transport matrix:
\[
j^* = \arg\max_{j} \hat{M}_{i,j},
\]
where $\hat{M} \in \mathbb{R}^{M_s \times M_t}$ is the normalized assignment matrix.
Matches assigned to the dustbin (i.e., $j^* = M_t + 1$) are discarded. The remaining matches are considered for geometric consistency.

\paragraph{Geometric Consistency.}
For each matched pair $(\omask_{s,i}, \omask_{t,j^*})$, we project both masks into 3D using the depth maps and camera poses to obtain their corresponding point clouds, $\mbsp_{s,i}$ and $\mbsp_{t,j^*}$. To evaluate geometric consistency, we compute the geometric similarity between the two point-clouds in the same way as done by Concept-Graphs. A match is accepted if $\geoSim(i, j^*) > 0.5$. All matches failing this geometric filter are rejected.

\paragraph{Evaluation on Replica.}  
We use the RGB-D frames from Semantic-NeRF~\cite{Zhi:etal:ICCV2021} and subsample each 900-frame sequence with a stride of 20, resulting in 45 frames per scene. We perform pairwise matching across all possible frame pairs (990 in total) and aggregate the correspondences to construct the final instance map. The corresponding results are reported in the third row of Table~\ref{tab:instance_mapping_replica_supplementary}.

\section{Additional Details on Object-Relative Topological Navigation}
\textbf{Background:} We used RoboHop~\cite{RoboHop} for benchmarking on the topological navigation task. RoboHop uses a segment-based topological map to enable open-world visual navigation. It extracts semantically-meaningful image segments using SAM~\cite{kirillov2023segment}. These segments become the nodes of a topological graph, where edges are formed both intra-image--linking segments within the same image based on pixel-level proximity--and inter-image--connecting corresponding segments across consecutive frames. For given query image segments, segment matching is performed using SuperPoint+LightGlue to retrieve matching segment nodes from the map, each of which has a precomputed path length to the given goal. Using these path lengths along with the pixel centers of the image segments, \textit{yaw} ($\psi$) is computed as below~\cite{podgorski2025tango,RoboHop}:
\begin{equation}
    \psi = \frac{K}{W} \sum_i w_i(x_i - c)
    \label{eq:robohop}; \quad w_i = \frac{e^{\tau l_i}}{\sum_i{e^{\tau p_i}}}
\end{equation}
where $c$ is the image center, $x_i$ is the segment center, $p_i$ is the path length, $\tau$ is the temperature parameter (set to $5$), $w_i$ is the softmax weight per query segment, $W$ is the image width, and $K$ is the proportional gain (set to $0.4$). For forward translation, a fixed velocity of $0.5m/s$ is used. We obtained the source code from the original authors of RoboHop~\cite{RoboHop}, and followed~\cite{podgorski2025tango} to evaluate using HM3Dv0.2~\cite{ramakrishnan2021hm3d}. This benchmark data is based on the validation set of the InstanceImageNav (IIN) challenge~\cite{krantz2022instance}, where $108$ episodes ($3$ each from the $36$ unique environments) are used to define the start position, object goal, and map trajectory. A topological graph is constructed a priori from the map trajectory, which is then used during the execution phase for localization, planning, and control.

\begin{figure}[!htbp]
\centering
    \includegraphics[width=0.8\linewidth,
                    keepaspectratio]{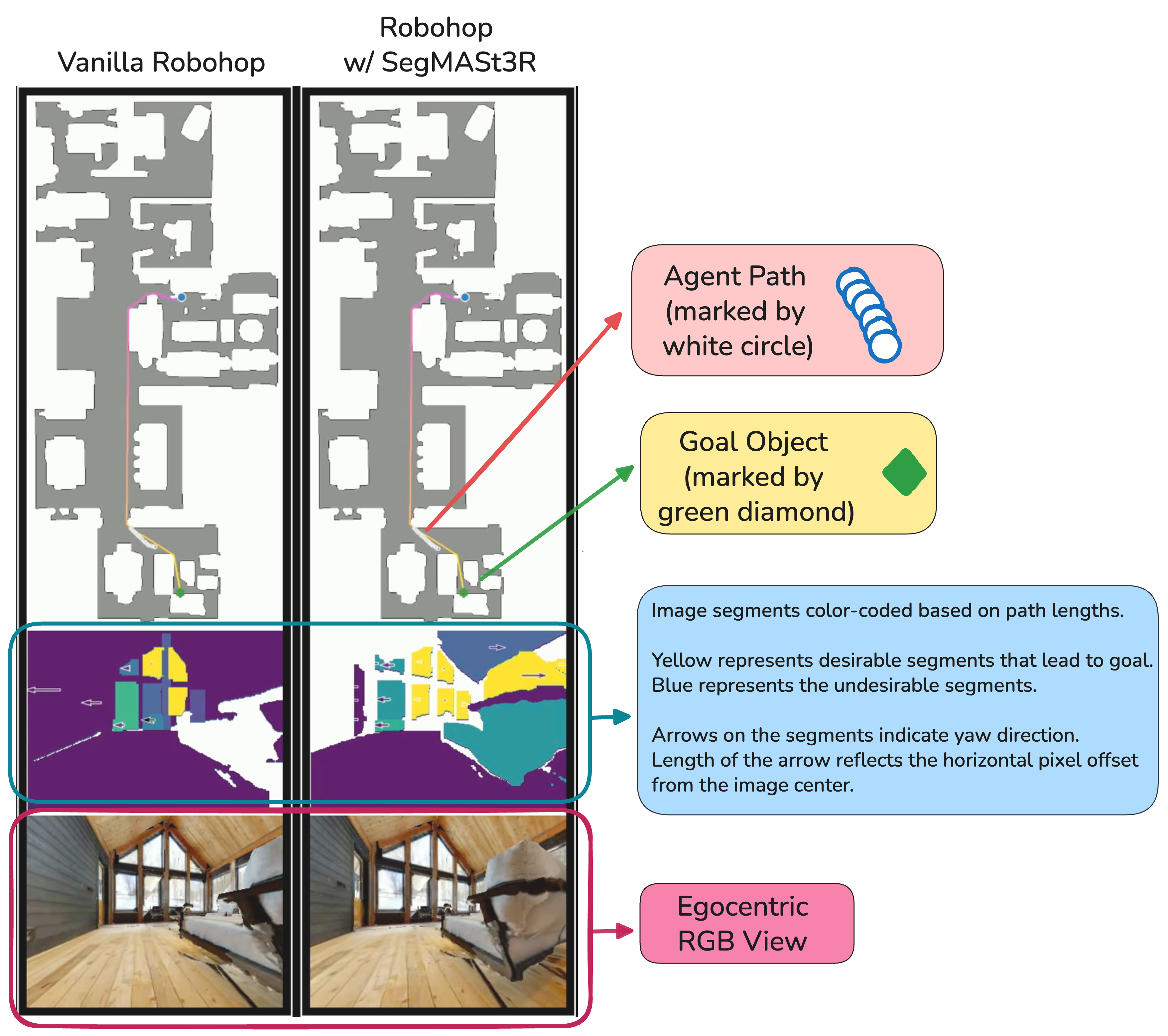}
    \caption{Different components of the image based goal navigation video.}
    \label{fig:navigation-legend}
\end{figure}

\textbf{Qualitative Visualizations:} In our experiments, we replaced RoboHop's segment matching during localization with \oursMSF{}, and compared vanilla RoboHop with this  version. In this supplementary material, we include navigation videos that qualitatively compare the two methods. The left half of the video corresponds to vanilla RoboHop, while the right half corresponds to our \oursMSF{}-enhanced version. Here, we provide an overview of the visualization panel of the video. At the top, we show an overhead view of the simulator, where the map trajectory is displayed with a color gradient starting from the blue circle and ending at the green diamond (goal object). The agent is initialized $5m$ away from the goal along the map trajectory, and its current state is displayed with a white circle. In the middle, we show the image segments which are color-coded based on the path lengths (normalized per image): yellow represents desirable segments (high weight $w_i$) that lead to the goal and blue represents the undesirable segments. The arrows on the segments indicate yaw direction ($\psi$), and the length of the arrow reflects its horizontal pixel offset from the image center ($(x_i - c)$). At the bottom, we show the egocentric RGB view from the agent's current state. 

We compare our method against the vanilla RoboHop baseline on four HM3D episodes via side-by-side video clips, emphasizing two key aspects:

\begin{enumerate}
  \item \textbf{Success vs.\ Failure}:  
    \texttt{nav\_bed\_20.mp4} and \texttt{nav\_chair\_55.mp4} demonstrate that, with our segment-matching module, the agent successfully reaches the goal, whereas with the vanilla RoboHop method it fails to reach the goal.
  \item \textbf{Fewer Steps}:  
    \texttt{nav\_sofa\_9.mp4} and \texttt{nav\_chair\_8.mp4} show that our model not only completes the task reliably but does so in fewer steps (around 40 steps in the former episode), while the baseline occasionally hesitates or deviates for a short interval before recovering.
\end{enumerate}

\section{Checklist Justifications}

\subsection{Limitations of Proposed Method}
\begin{itemize}
    \item \textbf{Reliance on 3D Instance-Segmentation Annotations}: Training is straightforward when 3D instance-level segmentation ground truth is available, as in datasets like ScanNet++ and Replica. For datasets without such annotations such as MapFree, our method can be trained using pseudo-ground truth generated via segment tracking approaches such as SAM2~\cite{ravi2024sam}. However, tracking errors and inconsistencies in pseudo-labels may propagate through training and impact final matching performance. Investigating robust loss functions, confidence weighting, or semi-supervised training schemes to better handle imperfect annotations remains an important direction for broader applicability.
\end{itemize}

\subsection{Code of Ethics}
\begin{itemize}
    \item \textbf{Research involving human subjects or participants}: We work on publicly available datasets that have been cited. No human subjects or their participation is involved.

    \item \textbf{Data-related concerns}: We utilized commonly used datasets without any human beings in them. So privacy is not a concern. The datasets are publicly available and have been cited. We use them in accordance with their respective licenses. 

    \item \textbf{Societal Concerns and Broader Impact}: While the work poses no immediate societal risks, its automation potential could influence future labor markets. Conversely, it promises clear benefits for a range of indoor robot-navigation tasks. 

    \item \textbf{Impact Mitigation Measures}: We will make the code, trained models and research artifacts publicly available to allow for reproducibility and mitigation in case unforeseen negative consequences arise from this research.

\end{itemize}

\subsection{Compute Resources}
\label{subsec:broader-impact}
Our approach fine-tunes an existing foundation model, keeping computational demands and therefore its carbon footprint low. Our largest proposed model requires only a \textbf{single RTX A6000 GPU and 22 hours of training}. Using a CO$_2$-to-power ratio of  0.7 kg CO$_2$e / kWh factor, one training run emits $\approx$ 4.8 kg CO$_2$e.

\end{document}